\documentclass[journal,comsoc]{IEEEtran}
\usepackage[T1]{fontenc}
\usepackage{epsfig}
\usepackage{amsmath,bm}
\usepackage{algorithm}
\usepackage{algorithmic}
\usepackage{subfigure}
\usepackage{epstopdf}
\usepackage{enumerate}
\usepackage{tabularx}
\usepackage{amsthm}
\usepackage{booktabs}
\usepackage{balance}
\usepackage{xspace}
\usepackage{url}

\newcommand{\ourmethod}{\texttt{CST-GL}\xspace}
\newcommand{\mts}{multivariate time series\xspace}
\newcommand{\mtsad}{multivariate time series anomaly detection\xspace}
\newcommand{\revision}{\textcolor{black}}
\newcommand{\minorrevision}{\textcolor{black}}

\usepackage{soul}
\usepackage{xcolor}

\ifCLASSINFOpdf
\else
\fi
\usepackage{comment}

\usepackage{amsmath}
\usepackage{commath}
\usepackage[cmintegrals]{newtxmath}
\begin{document}
%

\title{Correlation-aware Spatial-Temporal Graph Learning for Multivariate Time-series Anomaly Detection}
%
%
%

\author{Yu Zheng, Huan Yee Koh, Ming Jin, 
Lianhua Chi, Khoa T. Phan, Shirui Pan, Yi-Ping Phoebe Chen, Wei Xiang
\thanks{Y. Zheng, L. Chi, K. T. Phan, Y-P. P. Chen, and W. Xiang are with Department of Computer Science and Information Technology, La Trobe University, Melbourne Australia. 
E-mail: \{yu.zheng, l.chi,  k.phan, phoebe.chen, w.xiang\}@latrobe.edu.au. 
}
\thanks{H. Y. Koh and M. Jin are with the Department of Data Science and AI, Faculty of IT, Monash University, Clayton, VIC 3800, Australia. 
E-mail: \{ming.jin, huan.koh\}@monash.edu. 
}
\thanks{S. Pan is with School of Information and Communication Technology, Griffith University, Australia. 
Email: s.pan@griffith.edu.au. 
}
\thanks {Y. Zheng and H. Y. Koh contributed equally to this work.}
\thanks {L. Chi is the corresponding author.}

\thanks{Manuscript received Jan 3, 2022; revised xx xx, 202x.}
}

%
%

\markboth{Journal of \LaTeX\ Class Files,~Vol.~x, No.~x, August~2020}%
{Shell \MakeLowercase{\textit{et al.}}: Bare Demo of IEEEtran.cls for IEEE Communications Society Journals}
%



\maketitle

\begin{abstract}
Multivariate time-series anomaly detection is critically important in many applications, including retail, transportation, power grid, and water treatment plants. Existing approaches for this problem mostly employ either statistical models which cannot capture the non-linear relations well or conventional deep learning models (e.g., CNN and LSTM) that do not explicitly learn the pairwise correlations among variables. To overcome these limitations, we propose a novel method, correlation-aware spatial-temporal graph learning (termed \ourmethod), for time-series anomaly detection. \ourmethod explicitly captures the pairwise correlations via a \mts correlation learning module based on which a spatial-temporal graph neural network (STGNN) can be developed. Then, by employing a graph convolution network that exploits one- and multi-hop neighbor information, our STGNN component can encode rich spatial information from complex pairwise dependencies between variables. With a temporal module that consists of dilated convolutional functions, the STGNN can further capture long-range dependence over time. A novel anomaly scoring component is further integrated into \ourmethod to estimate the degree of an anomaly in a purely unsupervised manner. Experimental results demonstrate that \ourmethod can detect and diagnose anomalies effectively in general settings as well as enable early detection across different time delays. Our code is available: \url{https://github.com/huankoh/CST-GL}
\end{abstract}



%
\IEEEpeerreviewmaketitle

\section{Introduction}\label{sec:intro}

\IEEEPARstart{R}{apid} developments in Cyber-Physical Systems (CPS) have resulted in an explosive growth of time-series data collected across industries. In many applications, the CPS implemented generates time-series data from multiple devices or sensors, forming a complex \textit{multivariate time-series}. Importantly, an operator may have thousands to millions of CPS systems, recording a manually unmanageable amount of multivariate time-series data. For example, each server of a cloud infrastructure provider generates multivariate time-series data and many providers may have up to over millions of servers \cite{li2021multivariate}. A similar scale in the CPS system has also been observed in numerous commercial systems and critical infrastructures including power systems, spacecraft \cite{hundman2018detecting}, engines, transportation, cyber networks \cite{su2019robust}, and water treatment plants \cite{mathur2016swat}. Relying on human labours to monitor these operations would thus be not only impractical but also impossible. 

To enable effective monitoring and warning of large-scale system operations, multivariate time-series anomaly detection has become an important topic. Successful implementation of multivariate time-series anomaly detection model could bring substantial economic and social benefits. For instance, in a water treatment plant \cite{ahmed2017wadi}, hundreds of sensors are installed to monitor water level, flow rates and water quality. A malicious attack may occur by simply turning on a single motorized valve, causing a disastrous cascading effect on the entire water distribution system. Automatically monitoring and detecting these abnormal behaviours can thus provide a fast response, which helps rectify errors, reduce cost, and save lives.  

Among the various implementation approaches for detecting anomaly events, \textit{unsupervised} anomaly detection is one of which has attracted the most attention due to the difficulty of obtaining ground-truth anomalies over time. Early approaches typically employed either statistical unsupervised models such as ARIMA/VAR \cite{yu2016improved} or distance-based approaches \cite{keogh2005hot,wang2018exact}. Unfortunately, these methods cannot capture the non-linear spatial and temporal relationship from the multivariate time-series data well. More recently, with the flourish of deep learning (DL), significant advances have been made. For instance, Hundman et al. proposed a Long Short-Term Memory (LSTM) network together with a nonparametric thresholding approach  \cite{hundman2018detecting} and Su et al. proposed a representation learning-based stochastic recurrent neural network \cite{su2019robust} approach to improving the current ability to detect multivariate time-series anomaly events. While the proposed DL frameworks can efficiently scale through high-dimensional multivariate time-series data, they did not explicitly model the underlying pairwise inter-dependence among variable pairs, weakening their capacity in detecting complex anomaly events. 

\begin{figure}[t]
    \centering
       \includegraphics[width=0.48\textwidth]{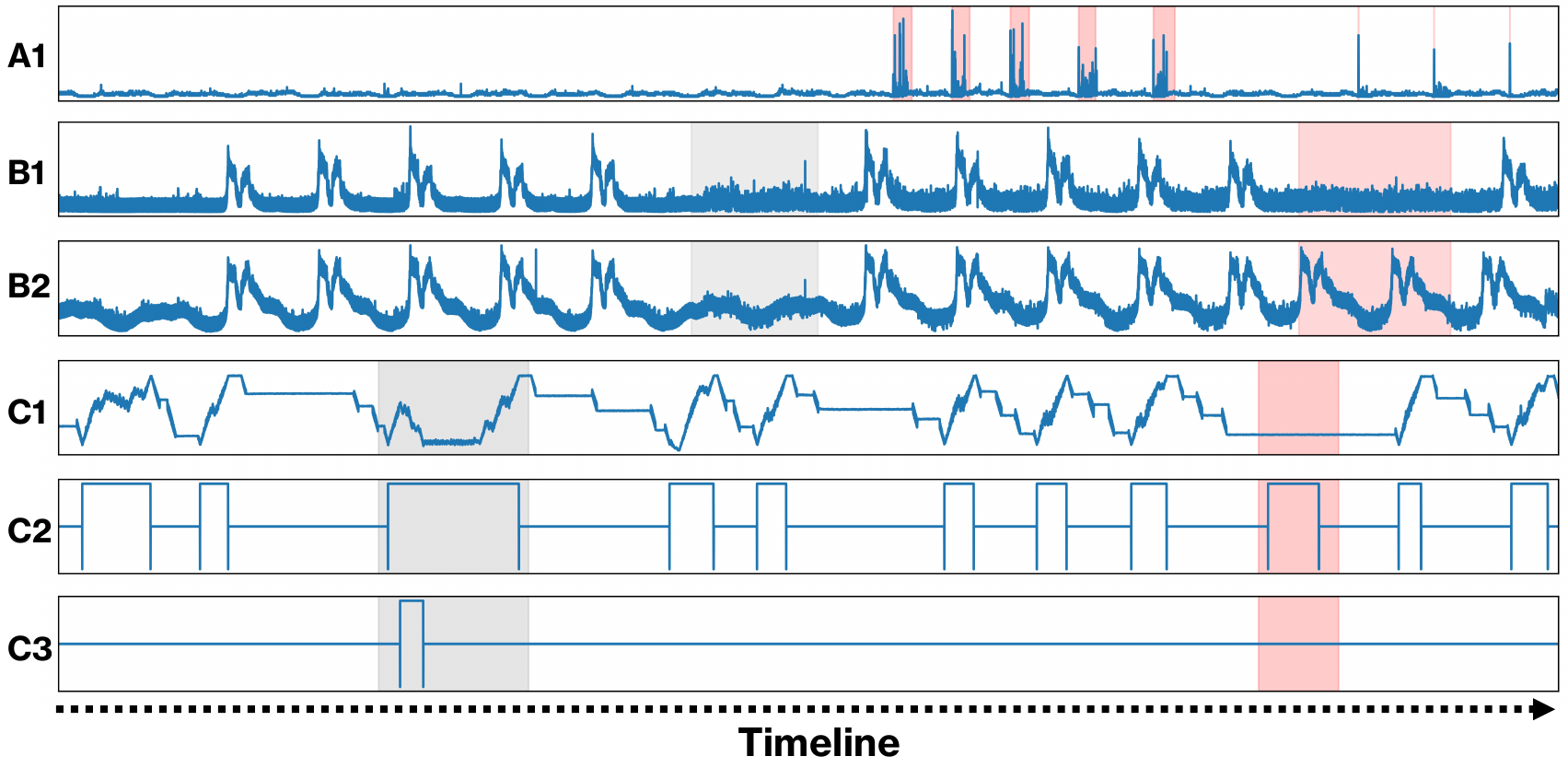}\\
       {\caption{An example of multivariate time-series data. Red span represents anomaly events while grey span represents time event highlights that are non-anomalous but form a reference to compare the behaviours of anomaly events. \revision{The first three examples showcased are from the Server Machine Dataset (SMD)~\cite{su2019robust}, which contains data from internet servers, while the last three are drawn from the WADI, a water treatment plant sensing dataset~\cite{ahmed2017wadi}.}}\label{fig. intro}}
       \vspace{-0.3cm}
\end{figure}

The difficulty of detecting anomaly events in multivariate time-series data lies in the fact that the variable pairs are intricately related. Figure \ref{fig. intro} shows a real-world inspired example of multivariate time-series data with six variables where A, B and C represents closely related variable groups. A1 variable is not closely related to other variables and the detection of anomaly events can simply be a significant deviation from past behaviours. B1 and B2 are two inter-related variables that should go up and down together, a deviation from this relationship is thus an anomaly event. On the other hand, C1 always increases with a lag after C2 is switched on (upward spike). The exception to the C1-C2 relationship (grey span of Figure \ref{fig. intro}) is when C3 is also switched on as C3 decreases C1, creating an offsetting effect on C2. The red span in the C variable groups indicates that an anomaly event has occurred because C1 does not increase despite C2 being switched on and C3 being switched off. While variable pairs that form the multivariate time-series are naturally interdependent, the degree of inter-dependence tells the full story of multivariate time-series data. Further, as shown above, the complexity increases exponentially with the increase in the number of variables. It is thus crucial for an anomaly detection model to not only assume the inter-dependent relationship but to \textit{explicitly learn and capture the pairwise correlations (i.e., degree of spatial dependence) between the variables of a multivariate time-series}. 

To explicitly capture the pairwise correlations, a natural way is to model \mts as a graph. For example, by treating each sensor as a node, a sensor graph can be constructed in which the node features are continuously changed over time. With a representative graph, spatial-temporal graph neural networks (STGNNs) can then be employed to tackle the \mtsad task by explicitly modeling the pairwise correlations via a graph neural network (GNN) module and temporal information via a CNN \cite{li2019spatio} or RNN \cite{seo2018structured} module. However, generic STGNN models necessitate a predefined graph to explicitly model pairwise correlations between variable pairs, which is often absent in many multivariate time-series datasets. \revision{Consequently, while previous STGNN methods are equipped to capture spatial dependencies, their capacity to learn and construct the relationship between the variables may not always be optimized, especially in cases where a predefined graph is not readily available in many multivariate time-series data.}

To address the limitations of generic STGNNs, \revision{Graph Deviation Network (GDN)} \cite{deng2021graph} employs a simple graph learning layer to learn and construct the pairwise correlation relationship between the variable pairs. Then, a graph attention network is used to propagate historical information among the variables to forecast the next observation. Anomaly events are subsequently detected based on the magnitude of deviation between forecast and real observations. Nevertheless, in this preliminary study, GDN only captures the spatial dependency in the direct neighbors of each variable, which may cause it to lose important information from high-order (multi-hop) neighbors. \revision{Furthermore, it did not explicitly model temporal relations within each univariate time series, which are crucial for characterizing \mts data~\cite{jin2022multivariate} and thus further compromises the effectiveness of GDN.}

Based on the above observations, we summarize the challenges for \mtsad from the graph neural network perspective as follows.
\begin{itemize}
\item \revision{\textbf{Multivariate time-series correlation learning (Challenge 1).}} The underlying correlations among the time-series variable pairs are important for \mtsad task. How to explicitly capture the pairwise relations among variables to enable spatial-temporal analysis is the first challenge.
\item \revision{\textbf{Spatial-temporal dependency modeling (Challenge 2).}} Multivariate time-series analysis requires a deep understanding of spatial-temporal dependency; how to simultaneously capture spatial and temporal dependency remains a challenge for \mts. 
\item \revision{\textbf{Anomaly scoring (Challenge 3).}} How to estimate the anomaly score in an unsupervised way is the ultimate challenge for \mtsad. 
\end{itemize}

To address these challenges, we propose a novel algorithm \ourmethod in this paper. Our theme is to model \mts as a graph and design a spatial-temporal graph neural network to perform forecasting. Based on the forecasting results, the anomaly score can be well estimated, and the anomalies can be detected accordingly. To be more specific, we propose a \mts correlation learning module that can automatically infer the underlying correlation among variables (\textit{for Challenge 1}). Then, a well-designed spatial-temporal graph neural network is presented to model both the spatial and temporal dependency (\textit{for Challenge 2}). \revision{The spatial dependence is modeled via a graph convolution network based on the gated mix-hop feature propagation that exploits neighbors from both single and multiple hops to better encode spatial information.} The temporal dependence is captured via a temporal convolutional network which incorporates a gating mechanism with convolution functions for long dependence modeling. Based on the forecast, we introduce a robust PCA-based anomaly scoring module to compute the anomaly scores, enabling effective anomaly detection and diagnosis (\textit{for Challenge 3}). Experimental results on real datasets demonstrate the superb performance of our method.

The main contributions of this paper are as follows:

\begin{itemize}
\item We propose an integrated algorithm for \mts data analysis. Our method seamlessly integrates correlation learning into a spatial-temporal network for \mts.
\item We propose a novel algorithm for \mtsad. Our method can automatically detect anomalies from complex time-series via auto-thresholding and enable early detections effectively. 
\item We compare our method with \revision{\textit{eleven}} baselines for both general \mtsad which aims to evaluate the overall performance in a whole dataset, as well as early detection of anomalies that needs to detect anomaly as early as possible. Our experimental results demonstrated that our method outperforms all baselines in both settings.
\item We conduct a case study to demonstrate that our method not only enables effective anomaly detection but also provides interpretability in real-life applications. 
\end{itemize}

The rest of the paper is structured as follows. Section~\ref{sec:rw} reviews the related work. Section~\ref{sec:definition} gives the definition of the task. Section~\ref{sec:model} presents the proposed \ourmethod. Section~\ref{sec:experiments} illustrates our experiments and conclusion in Section~\ref{sec:conclusion}.

\section{Related Work}\label{sec:rw}
In this section, we introduce the past work on multivariate time-series anomaly detection and graph neural networks.
\subsection{Anomaly Detection in Multivariate time-series}\label{subsec:rw_mts_ad}
Detecting anomalies in time-series is a challenging task that has been perennially studied \cite{chandola2009anomaly,chalapathy2019deep,blazquez2021review}. Historically, statistical models such as ARIMA/VAR  \cite{yu2016improved}, PCA \cite{li2014model} and SVM \cite{manevitz2001one} have been applied to detect anomalies in univariate and multivariate time-series. Traditional techniques involving wavelet analysis \cite{lu2008network}, non-parametric \cite{siffer2017anomaly}, pattern-based \cite{feremans2019pattern,yeh2016matrix} and distance-based \cite{keogh2005hot,wang2018exact} approaches have also been collaboratively implemented. More recently, substantial efforts have been made to advance deep learning approaches for anomaly detection in multivariate time-series data across numerous domains \cite{hundman2018detecting,su2019robust,zhang2019deep}. As argued by \cite{wu2020connecting,garg2021evaluation}, this phenomenon has arisen because (a) deep learning frameworks are free from stationary assumptions and can scale through high dimensional temporal data and (b) unlike pattern-based approaches that only detect anomaly events by identifying anomalous sub-sequences, deep learning frameworks can detect anomalous event timestamp-by-timestamp within sequences and are thus well suited for the deployment of real-time streaming anomaly detection systems. 

Deep learning models for multivariate time-series anomaly detection are primarily designed using recurrent neural network (RNN) that are combined either with convolutional neural networks (CNN) \cite{zhang2019deep,tayeh2022attention}, variational autoencoder (VAE) \cite{park2018multimodal,su2019robust} or Generative Adversarial Networks \cite{li2019mad}. The RNN is employed to capture temporal dependencies \cite{chauhan2015anomaly,malhotra2016lstm,hundman2018detecting} while the CNN, VAE or GAN is incorporated to capture dependencies among the multivariate variables. Any time-series observations which unexpectedly deviate from the learned temporal and relational dependencies would then be treated as anomalies. However, since CNN, VAE and GAN do not explicitly learn the relationship between the multivariate variables and only encapsulate interactions among variables into a global hidden state, they cannot fully exploit the latent dependencies between the variable pairs\revision{~\cite{chen2022deep,han2022learning}}. For more research on deep learning for time-series anomaly detection, we refer readers to the most recent survey \cite{darban2022deep}. 

\subsection{Graph Learning}\label{subsec:rw_gnn}
\revision{Graph learning~\cite{xia2021graph} is a new learning paradigm that enables machine learning for graph data. A key component of this paradigm, graph neural networks, have been widely studied to handle an array of graph-structured data~\cite{zhang2022trustworthy}. This includes a well-known subset of methods, namely spatial-temporal graph neural networks, which are typically applied to modeling multivariate time series~\cite{zhang2023self}. In this context, graph structure learning is often involved when prior knowledge of the underlying graph topology is not readily available.
}

\textit{Generic graph neural networks.} 
Graph neural networks (GNNs) have recently become de facto models to exploit graph data for graph analytics \cite{kipf2016semi,velivckovic2017graph,liu2022survey,zheng2022graph,jin2021multi,huang2022you}. The core idea of graph neural networks is to employ a \textit{message passing} scheme, which iteratively updates the representation (embedding) of a target node by propagating the representations of neighboring nodes. For instance, Graph Convolution Network (GCN) \cite{kipf2016semi} updates its node embedding by assigning a predefined weight to each message (embedding) from a neighbor. GAT \cite{velivckovic2017graph} automatically learns the weight of each neighbor and performs a weighted aggregation to update the target node's representation. Due to the capacity of modeling inter-relationship of different entities in various domains, GNNs have been widely used in domains and applications including traffic \cite{wu2021traversenet,zhang2023self}, recommender systems \cite{jin2022neural}, drug discovery \cite{nguyen2023gpcr,koh2023psichic}, and anomaly detection \cite{zheng2022unsupervised, zheng2021generative}.

\textit{Graph Structure Learning.} Learning graph neural network models typically requires a predefined graph structure so that the \textit{message passing} can be performed along with the topological structure. \revision{However, in many applications related to time series, the graph structure may be not available and the GNN models are not directly applicable.} To overcome this challenge, graph structure learning \cite{jin2020graph, liu2022towards} recently has emerged to automatically learn the graph structure from the data itself. For instance, SUBLIME \cite{liu2022towards} presents a structure bootstrapping contrastive learning framework to infer the relationship among data. However, these approaches can only be applied to static data. For dynamic data such as time series considered in this paper, these methods cannot be directly applied.

\textit{Spatial-Temporal Graph Neural Networks.} To extend GNNs for handling dynamic graph-structured data, recent research has delved into STGNNs \cite{jin2023survey}. These are especially effective in situations where the underlying graph structure remains static, but the features of the nodes undergo dynamic changes over time. A prime example of STGNNs in action is traffic forecasting, where the physical infrastructure such as subway stations and tracks is constant, but the traffic volume fluctuates continuously. Seo et al. \cite{seo2018structured} proposed a recurrent STGNN which adopts the Long-LSTMs \cite{hochreiter1997long} and GCN \cite{kipf2016semi} as key components to capture temporal and spatial dependencies. Instead, Li et al. \cite{li2019spatio} proposed a CNN-based method (1D convolution) to capture the capture temporal dependencies and a GCN to capture spatial dependencies. Wu et al. \cite{wu2020connecting} propose a joint graph structure learning and forecasting framework for spatial-temporal modeling. \minorrevision{The ability of STGNNs to handle dynamic graph-structured data has spurred interest in various domains, including traffic management \cite{pei2023selfemission}, infrastructure monitoring \cite{liang2023Social}, cloud computing \cite{he2023cloud}, human activity analysis \cite{yu2022human}, and anomaly detection \cite{deng2022gcad,huang2023videoanomaly}. While the aforementioned STGNN methods concentrate on graph learning and spatial-temporal pattern modeling, \ourmethod stands out by integrating two pivotal components: first, it incorporates a graph learning module with STGNN to accurately capture spatial-temporal dependencies; and second, it introduces a robust anomaly scoring module. This scoring module efficiently separates anomalous signals from the noise in \mts data forecasted by \ourmethod's STGNN module. By doing so, it maximizes the benefits of the relational structures learned by \ourmethod's graph learning module, facilitating effective anomaly detection and diagnosis.}

\section{Problem Formulation}\label{sec:definition}
A multivariate time-series with $T$ successive observations of equal-spaced samples as represented $\mathbf{X} =\{\mathbf{x}^\textit{1}, \mathbf{x}^\textit{2}, \cdots, \mathbf{x}^\textit{T}\},\mathbf{x}^t \in \mathbb{R}^{N}$ is composed of $N$ number of univariate time-series $\{\mathbf{x}^t_{1},\mathbf{x}^t_{2}, \cdots,\mathbf{x}^t_{N}\}$. In a real-time fashion, the multivariate time-series anomaly detection task requires learning of a scoring function, $A(\cdot)$, that outputs an anomaly score to current observation $T$ so that we have $A(\mathbf{x}^\textsc{a}) > A(\mathbf{x}^\textsc{m})$ where $\mathbf{x}^{\textsc{a}}$ is anomalous observation and $\mathbf{x}^{\textsc{m}}$ is not. Ideally, the proposed framework should also output a binary label that indicates whether a timestamp is anomalous or not, where $y^T \in \{0,1\}$ and $y^T = 1$ if the observation $\mathbf{x}^T$ is anomalous. In this paper, we tackle unsupervised \textit{real-time} anomaly detection. A model first learns normality from a non-anomalous training set. Then, as it receives a stream of observations containing both normal and anomalous data, it must detect anomalies in real-time, basing decisions solely on past data and cannot reverse its previous decisions.

\section{Methodology}\label{sec:model}
\begin{figure*}[!hbt]
    \centering
       \includegraphics[width=0.95\textwidth]{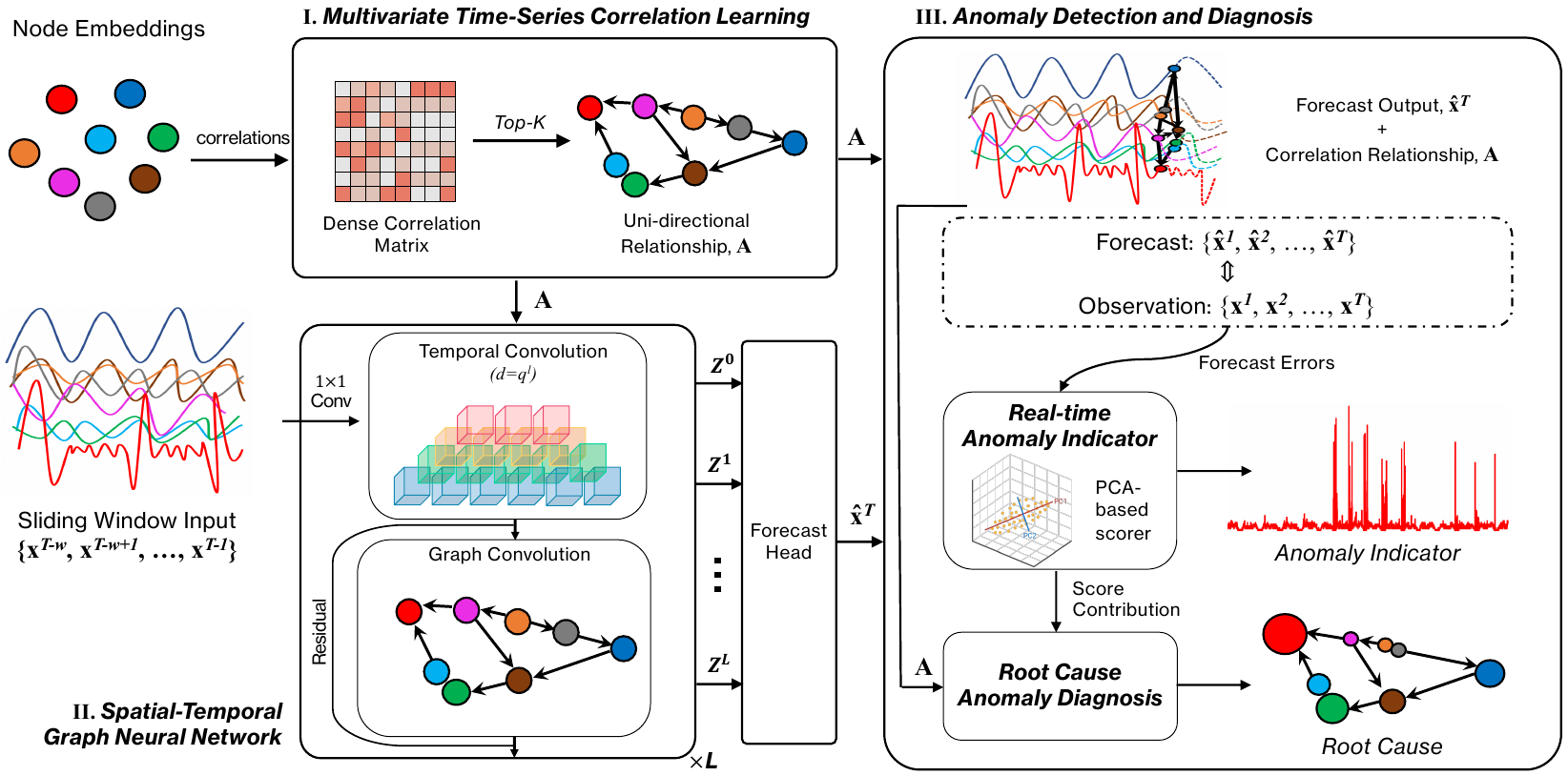}\\
       {\caption{Overall Framework of \ourmethod. \textbf{I. MTCL} starts with a randomly initialized node embedding for each multivariate variable, and  learn the underlying graph adjacency matrix, $A$, adaptatively with the entire model in an end-to-end manner. The adjacency matrix, $A$, will be used by the graph convolution networks in the STGNN module. \textbf{II. STGNN}'s $1\times1$ convolution layer projects the sliding window input into the latent space. Then, the temporal and graph convolution networks are interlaced to capture rich spatial and temporal dependencies, representing one spatial. \revision{Skip connections, $Z^0 + Z^1 + ... + Z^L$, are incorporated to obtain hidden features that encapsulated the spatial-temporal patterns}. Finally, the forecast head module projects the hidden features into a one-step forecast output, $\hat{x}^T$. \textbf{III. ADD.} a) Real-time Anomaly Indicator: takes in the current one-step forecast result and all observation-forecast pairs computed prior to timestamp $T$, computes normalized forecast deviation and outputs an anomaly indicator score in real-time using PCA-based scorer. b) Root Cause Anomaly Diagnosis: takes in result from Real-time Anomaly Indicator and learned pairwise correlation, $A$, from MTCL to enhance \ourmethod's interpretability and identify the root causes of anomaly events.}\label{fig. framework}}
\end{figure*}
In this section, we present the overall framework of \ourmethod and its detailed designs to detect anomaly events in a multivariate time-series. As shown in Figure \ref{fig. framework}, our method mainly consists of three main constituents: I. \textit{multivariate time-series correlation learning}, II. \textit{spatial-temporal graph neural network}, and III. \textit{anomaly detection and diagnosis} module. 

Given a multivariate time-series, we first propose to exploit the latent associations (i.e., edges) between each univariate time-series (i.e., nodes) explicitly via a pairwise correlation learner, where the learned graph structure together with the historical observations are then encoded by a sandwich-structured spatial-temporal graph neural network to make reliable forecasting. Specifically, we interlace the designed graph and temporal convolutions to capture rich spatial and temporal dependencies respectively. The underlying considerations are two-fold: (1) The potential anomalies in a univariate time-series can be easily identified by referring to its historical observations. For example, a sudden high CPU wattage is likely to trigger the system alert if compared with long-term historical readings. (2) However, for multivariate time-series data, the anomalies in a specific variable may not only associate with its historical observations but also the readings of other variables. A concrete example is traffic networks, where the change of road conditions in a street may cause a serious traffic jam in another one. Thus, it is crucial to model the underlying spatial and temporal dependencies in historical observations to perform precise and stable anomaly detection at each time step. To accomplish this goal, we propose a \textit{anomaly detection and diagnosis} module on the top of \textit{multivariate time-series correlation learning} and \textit{spatial-temporal graph neural network}, where the anomaly score at each time point is derived from the forecasting errors. In other words, we conjecture that time-series anomalies are typically reflected as the mismatch between \textit{anomalous observations} and the forecasting results given by the well-trained spatial-temporal model on \textit{non-anomalous data}.

Further, we argue that for root cause of anomaly events to be identified, pairwise correlations of variables have to be learned and captured by a proposed model. This is because univariates that deviate significantly from past spatial and temporal behaviours may only be symptoms of the root cause and the relationship of pairwise correlation can reveal the root cause variables. As shown later in the experimental section, \ourmethod identifies root cause of an anomaly events using the well-learned pairwise correlation graph that captures well the inter-dependence relationship between the variable pairs.

In the rest of this section, we introduce the multivariate time-series correlation learning (MTCL) in Subsection \ref{subsec:clm}. Then, in Subsection \ref{subsec:sgcn} and \ref{subsec:tcn}, we illustrate the detailed designs of the proposed spatial-temporal graph neural network (STGNN) in capturing the underlying spatial and temporal clues for accurate forecasting. Finally, we discuss how the proposed anomaly detection and diagnosis module can compute the real-time anomaly score in Subsection \ref{subsubsec:realtime_anomaly} and identify the root cause of an anomaly event in Subsection \ref{subsubsec:rootcause_anomaly}. 

\subsection{Multivariate Time-Series Correlation Learning}\label{subsec:clm}
To explicitly enable the modeling of pairwise dependencies among variables in a multivariate time series, we design a correlation learning layer and propose to learn the underlying unknown graph adjacency matrix $\mathbf{A}$ adaptively, where nodes and edges denote variables and their connectivity. Specifically, our detailed formulation is given as follows:
\begin{equation}
\left\{
\begin{aligned}
    & \widetilde{\mathbf{N}}_{1} = tanh(\alpha\mathbf{N}_{1}\mathbf{W}_{1}), \\ 
    & \widetilde{\mathbf{N}}_{2} = tanh(\alpha\mathbf{N}_{2}\mathbf{W}_{2}), \\ 
    & \mathbf{A} =ReLU\Big(tanh\big(\alpha(\widetilde{\mathbf{N}}_{1}\widetilde{\mathbf{N}}_{2}^\mathsf{T} - \widetilde{\mathbf{N}}_{2}\widetilde{\mathbf{N}}_{1}^\mathsf{T})\big)\Big),
\end{aligned}
\right.
\label{eq: graph construction}
\end{equation}
where $\mathbf{N}_{1}, \mathbf{N}_{2} \in \mathbb{R}^{N \times d}$ are two randomly initialized node embedding matrices, and $\mathbf{W}_{1}, \mathbf{W}_{2} \in \mathbb{R}^{d \times d}$ are two set of trainable parameters. The hyper-parameter $\alpha$ denotes the nonlinear activation saturation rate. Compared with our approach, many existing works construct such adjacency matrix by measuring the pairwise distance or similarity between variables in a multivariate time-series, such as Euclidean distance \cite{li2018diffusion} and Cosine similarity \cite{li2022mcgnet}, resulting in high time and space complexity of $\mathcal{O}(N^2)$ \cite{wu2020connecting}. Another significant drawback of existing methods based on distance or similarity metrics is that the learnt pairwise dependencies are symmetric, \revision{which is not desired in describing the relations between variables in real-world multivariate time series.} For example, the traffic jam on a street may cause the jam on another street but may not vice versa if there are alternative routes. 
\revision{Thus, we expect the learned time series dependencies to be uni-directional. Let $\widetilde{\mathbf{N}}_{1}$ and $\widetilde{\mathbf{N}}_{2}$ be two transformed node embedding matrices, the uni-directional property can then be achieved by the subtraction term $\widetilde{\mathbf{N}}_{1}\widetilde{\mathbf{N}}_{2}^\mathsf{T} - \widetilde{\mathbf{N}}_{2}\widetilde{\mathbf{N}}$ and two nonlinear activation functions, i.e., if $A_{ij}$ is a positive number, then $A_{ji}$ will be zero. The output adjacency matrix $\mathbf{A}$ will have all its elements regularized between 0 to 1.
To reduce the required computational cost and ease the optimization, we further mask elements with zeros in the learned graph adjacency matrix only except for the top-\textit{k} closest neighbors of each node to make $\mathbf{A}$ sparse controlled by the hyper-parameter $k$. Specifically, for $i$-th row in $\mathbf{A}$, we have the following post-processing:}
\begin{equation}
\left\{
\begin{aligned}
    & \mathbf{topk} = argmax(\mathbf{A}[i,:], k), \\ 
    & \mathbf{A}[i,-\mathbf{topk}] = 0, \\ 
\end{aligned}
\right.
\label{eq: graph construction post-processing}
\end{equation}
where $argmax(\cdot, k)$ returns the indices of top-$k$ largest values in the input vector.

\subsection{Spatial-Temporal Graph Neural Network}
\subsubsection{Graph Convolution Network}\label{subsec:sgcn}
The spatial correlations between variables play a vital role in reflecting the intrinsic dynamics of multivariate time-series. Towards this, we design a spatial graph convolution layer to effectively pass messages between variables and their neighbors to exploit the underlying spatial patterns, allowing better encoding historical observations to make more precise and stable predictions, thus benefiting the downstream anomaly detection tasks. 
Similar to SGC \cite{wu2019simplifying} and the discrete of MTGODE \cite{jin2022multivariate}, given an adjacency matrix $\mathbf{A}$ and the input (initial) states $\mathbf{H}_{in}$, we may characterize the graph propagation process as a combination of the feature propagation and linear transformation steps:
\begin{equation}
\left\{
\begin{aligned}
        & \mathbf{H}^{k+1} = \widetilde{\mathbf{A}} \ \mathbf{H}^{k},\  k \in \{0, \cdots, K\}, \\  
        & \mathbf{H}_{out} = \mathbf{H}^{K} \ \mathbf{\Theta},
\end{aligned}
\right.
\label{eq: SGC}
\end{equation}
where $K$ denotes the graph propagation depth, and we have $\mathbf{H}^{0}=\mathbf{H}_{in}$. Specifically, $\widetilde{\mathbf{A}}$ in the above equation denotes the normalized adjacency matrix, i.e.,  $\widetilde{\mathbf{A}}=\widetilde{\mathbf{D}}^{-1}(\mathbf{A}+\mathbf{I})$ and $\widetilde{\mathbf{D}}_{ii}=1+\sum_{j}\mathbf{A}_{ij}$. 

\revision{However, Equation \ref{eq: SGC} suffers from two critical limitations. Firstly, although the above feature propagation design allows to recursively propagate latent node states along with a given graph structure, it is inevitable to see that node latent states become indistinguishable, i.e., converge to a single point, or known as over-smoothing, with an increase of the propagation depth $K$ \cite{jin2022multivariate}. Secondly, only applying the linear transformation on the last node latent states $\mathbf{H}^{K}$ may be prone to errors~\cite{abu2019mixhop, wu2020connecting}. For example, if there are no correlations between variables in a multivariate time series, the feature propagation step will introduce noises to latent node states by blindly aggregating the neighbouring information. Thus, merely considering the linear transformation of the last propagated states hinders accurately modeling the latent spatial dynamics of a multivariate time series. To address these two limitations, we equipped the vanilla feature propagation in Equation \ref{eq: SGC} with a gating mechanism and replaced the upcoming linear mapping with an attentive transformation that mixes the information from multiple hops. We have the proposed graph convolution network defined as follows:}
\begin{equation}
\left\{
\begin{aligned}
        & \mathbf{H}^{k+1} = \beta \ \mathbf{H}_{in} + (1-\beta) \ \widetilde{\mathbf{A}} \ \mathbf{H}^{k},\  k \in \{0, \cdots, K\}, \\  
        & \mathbf{H}_{out} = \sum_{k=0}^{K}\mathbf{H}^{k} \ \mathbf{\Theta}^{k},
\end{aligned}
\right.
\label{eq: spatial convolution}
\end{equation}
where $\beta$ controls to retain how much original node information to avoid the aforementioned over-smoothing issue. Regarding the attentive transformation in Equation \ref{eq: spatial convolution}, we can easily alleviate the problem mentioned in the above example by assigning a relevant large weight to the initial node states $\mathbf{H}^{0}$ and small weights to $\mathbf{H}^{k}$, where $k \in \{1, \cdots, K\}$.

As mentioned in Subsection \ref{subsec:clm}, the learned pairwise dependencies are uni-directional. Thus, we refactor the final output of the graph convolution network as the summation of two transformations described in Equation \ref{eq: spatial convolution}, where the input latent node states are both $\mathbf{H}_{in}$ but with different adjacency matrices, i.e., $\mathbf{A}$ and $\mathbf{A}^\mathsf{T}$, to incorporate nodes' inflow and outflow information, respectively.

\subsubsection{Temporal Convolution Network}\label{subsec:tcn}
Solving Equation \ref{eq: spatial convolution} only allows to model the spatial dynamics at a certain point of time, where the rich temporal clues in multivariate time-series are neglected. To complete this missing information, we devise a simple yet effective temporal convolution network together with our graph convolution network to capture the expressive spatial and temporal patterns in historical observations.

We first introduce the composition of the proposed temporal convolution network, which consists of multiple residual dilated temporal convolution layers to extract and aggregate high-level temporal features in a non-recursive manner to avoid the shortcomings of Recurrent Neural Networks (RNNs), such as time-consuming iteration and gradient explosion \cite{jin2022multivariate, wu2020connecting, wu2019graph}. Specifically, given a sequence of historical observations $\mathbf{X} =\{\mathbf{x}^\textit{1}, \mathbf{x}^\textit{2}, \cdots, \mathbf{x}^\textit{T\revision{-1}}\}$, we have the a temporal convolution layer defined as follows:
\begin{equation}
    \mathbf{Z}^{l+1} = \mathcal{T}(\mathbf{Z}^{l}, Q^{l+1}) + TCN(\mathbf{Z}^{l}, {\mathbf{\Phi}}^{l}), \ l \in \{0, \cdots, L\},
\label{eq: temporal convolution layer}
\end{equation}
where the outputs of network is $\mathbf{Z}_{out}=\mathbf{Z}^L$, the input states $\mathbf{Z}^{0}$ are obtained by applying a linear mapping on $\mathbf{X}$, $TCN(\cdot, \mathbf{\Phi}^{l})$ is an temporal convolution function parameterized by $\mathbf{\Phi}^{l}$ at the $l$-th layer, and $\mathcal{T}(\mathbf{Z}^{l}, Q^{l+1})$ denotes a truncate function that taking the last $Q^{l+1}$ elements from $\mathbf{Z}^{l}$ along its sequence length axis. The underlying consideration is that the residual input $\mathbf{Z}^{l}$ has to be truncated to the length of $TCN(\mathbf{Z}^{l}, {\mathbf{\Phi}}^{l})$ before adding them together because the sequence length of latent node states shrinks gradually as the underlying temporal information is aggregated after each temporal convolution layers. Specifically, we have $Q^{l+1} = Q^{l} - r^l \times (k-1)$ and $Q^1=R-k+1$, where $k$, $r$ and $R$ are kernel size, dilation factor, and receptive field (i.e., $R=L(k-1)+1$ and $R=1+(k-1)(r^L-1)/(r-1)$ when $r=1$ and $r>1$). 
In terms of the design of temporal convolution function $TCN(\cdot, \mathbf{\Phi}^{l})$, we follow \cite{wu2020connecting} and adopt a gating mechanism to guide the information flow during the aggregation:
\begin{equation}
    \operatorname{TCN}(\mathbf{Z}^l, \mathbf{\Phi}^l) = f_{\mathcal{C}}(\mathbf{Z}^l, \mathbf{\Phi}^l_c) \odot f_{\mathcal{G}}(\mathbf{Z}^l, \mathbf{\Phi}^l_g),
    \label{eq: temporal convolution block}
\end{equation}
where $f_{\mathcal{C}}(\cdot)$ and $f_{\mathcal{G}}(\cdot)$ are filtering and gating convolutions, and $\odot$ denotes the element-wise product. Specifically, we define these two convolutions in below:
\begin{equation}
\left\{
\begin{aligned}
    & f_{\mathcal{C}}(\mathbf{Z}^l, \mathbf{\Phi}^l_c) = tanh\big(\mathbf{W}^{1 \times n}_{\mathbf{\Phi}^l_{c}} \star_{\Delta} \mathbf{Z}^l \ + \ \mathbf{b}^{1 \times n}_{\mathbf{\Phi}^l_{c}} \big), \\
    & f_{\mathcal{G}}(\mathbf{Z}^l, \mathbf{\Phi}^l_g) = sigmoid\big(\mathbf{W}^{1 \times n}_{\mathbf{\Phi}^l_{g}} \star_{\Delta} \mathbf{Z}^l \ + \ \mathbf{b}^{1 \times n}_{\mathbf{\Phi}^l_{g}} \big).
\end{aligned}
\right.
    \label{eq: temporal filtering and gating}
\end{equation}
In the above equation, $\star_{\Delta}$ denotes the dilated convolution operation, where the dilation $\Delta=r^l$. Specifically, to allow the model exploring multi-granular temporal clues and inspired by \cite{wu2020connecting}, $f_{\mathcal{C}}(\mathbf{Z}^l, \mathbf{\Phi}^l_c)$ and $f_{\mathcal{G}}(\mathbf{Z}^l, \mathbf{\Phi}^l_g)$ consists of multiple convolution filters (e.g., $\mathbf{W}^{1 \times n}_{\mathbf{\Phi}^l_{c}}$ and $\mathbf{b}^{1 \times n}_{\mathbf{\Phi}^l_{c}}$) with width $n \in \{2,3,6,7\}$. Since most of multivariate time-series data has some intrinsic periods \cite{wu2020connecting, wu2019graph}, such as 7, 14, 24, 28, and 30, the combination of the aforementioned kernel widths allows these common periods to be fully covered.

To simultaneously model spatial and temporal dynamics of a sequence of historical observations in a multivariate time-series, we construct a spatial-temporal graph neural network by combining the proposed spatial and temporal convolution networks, where the temporal and spatial convolution layers are interlaced, as shown in Figure \ref{fig. framework}. More precisely, a layer of the proposed spatial-temporal graph neural network is defined as follows by combining Equation \ref{eq: spatial convolution} and \ref{eq: temporal convolution layer}:
\begin{equation}
    \mathbf{Z}^{l+1} = \mathcal{T}(\mathbf{Z}^{l}, Q^{l+1}) + GCN\big(TCN(\mathbf{Z}^{l}, \mathbf{\Phi}^{l}), \mathbf{\Theta}\big),
\label{eq: spatial-temporal convolution layer}
\end{equation}
where $GCN(\cdot, \mathbf{\Theta})$ and $TCN(\cdot, \mathbf{\Phi}^{l})$ are defined in Equation \ref{eq: spatial convolution} and \ref{eq: temporal convolution block}. Finally, we take the output states $\mathbf{Z}_{out}$ to make a single-step-ahead forecasting via a multi-layer perceptron, i.e., $\hat{\mathbf{x}}^{T}=MLP(\mathbf{Z}_{out}, \mathbf{W}_{mlp})$, which forms a critical evidence to detect anomalies in a multivariate time-series.

\subsection{Anomaly Detection and Diagnosis} \label{subsec:anoscore}

\subsubsection{Real-time Anomaly Indicator}\label{subsubsec:realtime_anomaly}
With an effective joint learning of spatial and temporal dependencies from the non-anomalous data, it is expected that the anomalous observations in the test set deviate significantly from the learned patterns. Accordingly, to detect anomalous multivariate observations, we first compute the normalized forecasting deviation for every univariate variable and take the sum of the reconstructed univariate deviations to be the anomalous score for each multivariate observation.

\revision{Univariate variables within a multivariate time-series often possess vastly different attributes and scales. Consequently, we independently normalize each univariate deviation to preclude any single variable from dominating the aggregate multivariate deviation value. For every univariate variable, $\mathbf{x}^{T}_{i}$, we compute the absolute forecasting error, given by $\mathbf{e}^T_i = \abs{\mathbf{x}^{T}_{i} - \mathbf{\hat{x}}^{T}_{i}}$, at the current timestamp $T$. This error is then normalized:
$$ \mathbf{\tilde{e}}^T_i = \frac{\mathbf{e}^T_i - \bm{\mu}^T_i}{\bm{\sigma}^T_i}$$}
where $\bm{\mu}^T_i$ and $\bm{\sigma}^T_i$ are the median and inter-quartile range (IQR) value across error values, $\{\mathbf{e}^{T-W_a}_i,\mathbf{e}^{T-W_a+1}_i,...,\mathbf{e}^T_i\}$ in a sliding window where $W_a$ represents the window length. Our normalization approach is an extension of \cite{deng2021graph} where we acquire the median and IQR values through a sliding window rather than from the entire test set observations. This modification allows us to detect anomalies in real-time as normalizing error at time $T$ only relies on past observations and does not require future information \revision{as in \cite{deng2021graph}.} 

\revision{After the normalization of each univariate variable, we obtain a multivariate normalized error vector, $\mathbf{\widetilde{E}}^T \in \mathbb{R}^{1 \times N}$, for the current timestamp. Although prior research suggests directly taking the summation~\cite{zhao2020multivariate} or maximum~\cite{deng2021graph} to summarize the error vector into a single anomaly score at the current timestamp, we propose to leverage Principal Component Analysis (PCA) as an intermediate step before aggregating the normalized errors into a final anomaly score.} 

\revision{In particular, after training the spatial-temporal graph neural network module, we compute the normalized errors in the validation set, $\mathbf{\widetilde{E}}_{v}$. We fit a PCA on the validation normalized errors by finding the validation mean vector $\mathbf{\overline{E}}_v = mean(\mathbf{\widetilde{E}}_{v})$, their covariance matrix, $\mathbf{C}_v=cov(\mathbf{\widetilde{E}}_{v})$, and the orthogonal eigenvectors, $\mathbf{U}$. $\mathbf{U}$ consists of $N$ orthogonal eigenvectors associated with the $N$ largest eigenvalues in the diagonal matrix, $\mathbf{\Lambda}$, decomposed from $\mathbf{C}_v=\mathbf{U}\mathbf{\Lambda}\mathbf{U}^{-1}$. With the fitted PCA, we reconstruct the normalized errors at current timestamp}:
\revision{\begin{equation}
\left\{
\begin{aligned}
&\mathbf{P} = (\widetilde{\mathbf{E}}^T - \mathbf{\overline{E}}_{v})\mathbf{U}^T \\
& \mathbf{\widetilde{P}}, \mathbf{\widetilde{U}} = \mathbf{P}[:,L], \mathbf{U}[:,L] \\
& \mathbf{\widetilde{E}}^T_{\textsc{pca}} = \mathbf{\widetilde{P}} \,\mathbf{\widetilde{U}}^T + \mathbf{\overline{E}}_{v} \\
\end{aligned}
\right.
\label{eq: pca}
\end{equation}}
\revision{In the above equation, we first apply zero-centering to the normalized error at the current timestamp, $\mathbf{\widetilde{E}}^T$, by subtracting the mean validation error, $\mathbf{\overline{E}}_v$. We then project these results using validation eigenvectors, $\mathbf{U}$. Secondly, we keep only the first $L$ principal components. Finally, we reconstruct the normalized errors, $\mathbf{\widetilde{E}}^T_{\textsc{pca}}$, using the reduced $L$ dimensions and revert the zero-centering deduction by adding the validation mean error. We set $L$ as the number of components necessary to achieve a symmetric mean absolute percentage error (sMAPE) of less than 10\% on the validation set.}

\revision{With the reconstructed normalized error, we compute the final anomaly score at current timestamp by taking the L1 distance between the denoised and original normalized errors as the final anomaly score:} 
\revision{\begin{equation}
    A(T) = \rVert\mathbf{\widetilde{E}}^T_{\textsc{pca}} - \mathbf{\widetilde{E}}^T \rVert_1 
\label{eq: anomaly_score}
\end{equation}}
\revision{The incorporation of PCA addresses the fundamental problem posed by anomalies: the anomalous node variables have the potential to introduce bias into the learned embeddings within a neural network module, inadvertently affecting the forecast across all dimensions. This effect, corroborated by previous research~\cite{xu2018unsupervised}, often leads to an unwarranted increase in forecast errors in variable nodes that are otherwise unaffected. Even in the absence of anomaly events, certain variable nodes may sporadically experience an upsurge in errors due to random fluctuations~\cite{deng2021graph}. Such fluctuations can set off a cascade of effects across all nodes, echoing the impact of an actual anomalous event and potentially resulting in false positives. This unintended effect contributes to the degradation of accuracy in anomaly detection and diagnosis. }

\revision{Previous methodologies have attempted to resolve this issue with the utilization of Markov Chain Monte Carlo (MCMC) imputation~\cite{li2021multivariate,rezende2014stochastic}. This approach, however, is inefficient. In contrast, we propose the application of PCA to resolve these issues. PCA can efficiently project the normalized errors at current timestamp, $\mathbf{\widetilde{E}}^T$, onto the principal components of the validation errors, and subsequently reconstruct them as, $\mathbf{\widetilde{E}}^T_\textsc{pca}$. This process effectively dampens common noise variations and pinpoints variables that contribute significantly to anomaly events.  This identification is made possible because the variables that cannot be accurately reconstructed are more likely the true contributors to the anomaly events, thereby offering a more accurate depiction of the anomaly event itself.}
 
\revision{Though PCA is capable of mitigating the inherent noise for more accurate representation, it is still the ability of STGNN to capture spatial-temporal patterns that holds the key to a comprehensive solution. The joint implementation of STGNN and PCA is instrumental in detecting and diagnosis anomalies, as we demonstrate in Section V: Experimental Study.}

Last but not least, an anomaly indicator that can well signify the abnormality of a timestamp observation helps in informing system operators and determining an appropriate threshold by human experts to classify and detect anomalies. Nevertheless, for industrial operations that involve over thousands of multivariate time-series with distinct attributes such as warehousing robots \cite{chen2020unsupervised}, this approach does not scale well. To automate the threshold selection process, we classify current observation in the test set as anomalous if $A(T)$ in test set observation exceed the maximum $A(t)$ of all observations in the validation set. This non-parametric approach relies on \ourmethod's ability in sufficiently capturing the spatial and temporal dependencies of a multivariate time-series data, so that any observations that exceeds the maximum anomaly score during normal time period (i.e., validation data) are in fact anomalies while those that do not exceed the maximum value are not anomalies. 
\subsubsection{Root Cause Anomaly Diagnosis}\label{subsubsec:rootcause_anomaly}

\revision{Since the final anomaly score is calculated as the linear combination of reconstruction errors, we can identify the root cause of anomalous events by ranking the univariate variables that contribute most significantly to the anomaly score. In practical scenarios, we determine the percentage contribution of each univariate variable to the final anomaly score. This approach would provide a more detailed perspective, allowing operators to more effectively identify the root cause of anomalous events.}

In some cases, the top ranked variables that most contribute to the anomaly score may not be the root causes but are merely the symptoms \cite{deng2021graph, garg2021evaluation}. When the top ranked contributors are identified not to be the root cause, we further search for the variables that are most related to the top ranked contributors by aggregating the anomaly contribution scores of one-hop distance neighbors:
\revision{\begin{equation}
R_i(T) = \sum_{j \in N(i)} A_i(T) 
\end{equation}
\revision{where $A_i(T)$} represents the absolute error for the univariate node $i$ as per Equation \ref{eq: anomaly_score}. $N(i)$ represents the neighbors of the univariate node $i$, which is based on the learned relation between variable pairs from the MTCL module.}

As demonstrated in our experiments in Section \ref{sec:experiments}, this two-pronged approach ensures the systematic identification and diagnosis of (a) variables that exhibit abnormal behavior, and (b) variables closely related to these abnormal variables, as potential root causes of an anomaly event. \revision{The choice between directly ranking the variables based on the error contribution or based on the one-hop distance neighbors will largely depend on the nature of the anomalies.}

\section{Experimental Study}\label{sec:experiments}
In this section, we conduct experiments to explore \ourmethod's capabilities by answering following questions: 
\begin{itemize} 
	\item \revision{\textbf{Overall Detection Performance.}}  Does our framework outperform baseline methods in the unsupervised, real-time anomaly detection task? How do the individual modules within \ourmethod each contribute specifically to its ability to achieve anomaly detection and diagnosis?
	\item \revision{\textbf{Early Detection Performance.}} Can \ourmethod be adapted and generalized to commercial systems where early detection of anomaly events is often paramount?
	\item \revision{\textbf{Interpretability \& Case Study.}} Would \ourmethod benefit system operators in detecting and diagnosing multivariate time-series anomaly events in an interpretable manner?
    \end{itemize}

\subsection{Experimental Settings} \label{subsec:exp_setting} 
Here, we detail our experimental setup, covering datasets, baseline methods, and parameter configurations.
\subsubsection{Datasets}
We evaluated our method on three widely-used benchmark datasets: SWaT, WADI, and SMD. The statistics are presented in Table \ref{table:dataset}:
\vspace{-2.5mm}
\begin{table}[!hbt]
	\centering
	\caption{The statistics of the datasets.}
	\begin{tabular}{@{}c|c|c|c|c@{}}
		\toprule
		\textbf{Dataset} &  \textbf{$\sharp$ channels} & \textbf{$\sharp$ train} & \textbf{$\sharp$ test} &\textbf{anomalies} \\ \midrule
		\textbf{SWaT}   & 51    & 47,515     & 44,986            & 11.97\%              \\
		\textbf{WADI}             & 127    & 118,795     & 17,275         & 5.99\%               \\
		\textbf{SMD}      &   38     & 304,168  & 304,174       & 5.84\%           \\ 
		\bottomrule
	\end{tabular}
	\label{table:dataset}
\end{table}
\begin{itemize} 
	\item \revision{\textbf{SWaT~\cite{mathur2016swat}} is} a scaled-down version of a real-world industrial water treatment plant initiated by Singapore’s Public Utility Board. The dataset comprises 7 days of normal operations (train data) and 4 days of attack scenarios (test data). The anomaly labels represent the attacks that are conducted at different intervals in the test set.
	\item \revision{\textbf{WADI~\cite{ahmed2017wadi}} is} an extension of the SWaT dataset with a larger number of water pipelines, storage, and treatment systems, representing a more complete and realistic water treatment dataset \cite{deng2021graph}. The train set of WADI is two weeks of normal operation while the test set is a 2 days attack scenario. Following the original author's implementation \cite{deng2021graph}, we removed the first 21,600 samples and down-sampled SWaT and WADI to one measurement every 10 seconds by taking the median values.
	\item \revision{\textbf{SMD~\cite{su2019robust}} is} a real-world server machine dataset collected by a large Internet company. SMD contains time-series data of servers, each with 38 multivariate variables. It is divided into train and test sets of equal size. \revision{The original SMD dataset did not have any preprocessing applied to remove servers experiencing concept drift. This was subsequently addressed by the original authors in \cite{li2021multivariate} to remove servers suffering from concept drift. Following the subsequent work}, the reported results in this section takes the average scores computed for the 12 servers that do not suffer from concept drift.  
\end{itemize}

\begin{table*}[!hbt]
	\small
	\centering
	\caption{Average AUC performance ($\pm$ standard deviation) of five experimental runs on three benchmark datasets. \\The best and second best performing method in each experiment is in bold and underlined respectively.}
	\scalebox{0.86}{\begin{tabular}{p{55 pt}<{\centering}|p{59 pt}<{\centering}p{59 pt}<{\centering}|p{59 pt}<{\centering}p{59 pt}<{\centering}|p{59 pt}<{\centering}p{59 pt}<{\centering}}     
	\toprule[1.0pt]

		  & \multicolumn{2}{c|}{\textbf{SWaT}} & \multicolumn{2}{c|}{\textbf{WADI}}  & \multicolumn{2}{c}{\textbf{SMD}}\\ 
		  & {ROC} & {PRC} & {ROC} & {PRC}  & {ROC} & {PRC}
		    \\ 
		\cmidrule{1-7}
        \revision{Raw Signal} & \revision{0.8218 (0.0000)} &\revision{0.5661 (0.0000)} & \revision{ \underline{0.6544 (0.0000)}} & \revision{0.1117 (0.0000)} & \revision{0.7295 (0.0000)}& \revision{0.1775 (0.0000)}\\
        PCA         & 0.8257 (0.0000) & \underline{0.7298 (0.0000)} & 0.5597 (0.0000) & 0.2731 (0.0000) & 0.6742 (0.0000) & 0.2189 (0.0000) \\
	    AutoEncoder     & 0.8311 (0.0088) & 0.7224 (0.0094) & 0.5291 (0.0285) & 0.2210 (0.0205) & 0.8270 (0.0008) & 0.4388 (0.0046) \\
		Kmeans      & 0.7391 (0.0000) & 0.2418 (0.0000) & 0.6030 (0.0000) & 0.1158 (0.0000) & 0.5855 (0.0000) & 0.1308 (0.0000) \\
		DAGMM        & 0.7219 (0.0473) & 0.2630 (0.0932) & 0.5375 (0.0388) & 0.1315 (0.0396) & 0.7489 (0.0111) & 0.2522 (0.0095) \\
		LSTM-VAE    & 0.8016 (0.0016) & 0.6936 (0.0047) & 0.5165 (0.0322) & 0.1486 (0.0476) & 0.7802 (0.0105) & 0.3153 (0.0306) \\
		OmniAnomaly     & 0.8256 (0.0211) & 0.7061 (0.0066) & 0.5520 (0.0092) & 0.2184 (0.0023) & 0.8265 (0.0114) & 0.4575 (0.0197) \\
		USAD    & 0.8213 (0.0056) & 0.7087 (0.0055) & 0.5535 (0.0103) & 0.1945 (0.0008) & 0.7888 (0.0077) & 0.4686 (0.0011) \\
        \revision{MTAD-GAT} & \revision{0.8261 (0.0040)} &\revision{0.7176 (0.0043)} & \revision{0.4119 (0.0259)} & \revision{0.0729 (0.0013)} & \revision{\underline{0.8576 (0.0035)}} & \revision{\underline{0.5057 (0.0082)}}  \\
        GDN  & 0.8124 (0.0177) & 0.7135 (0.0035)    & 0.4725 (0.0056) & 0.0521 (0.0070) & 0.8443 (0.0150) & 0.4684 (0.0142) \\
		InterFusion     & \underline{0.8409 (0.0132)} & 0.6970 (0.0844) & 0.6388 (0.0311) & \underline{0.3775 (0.0319)} & 0.8374 (0.0215) & 0.4265 (0.0351) \\
		\cmidrule{1-7}
		\ourmethod       & \textbf{0.8520 (0.0022)} &  \textbf{0.7628 (0.0032)} & \textbf{0.8283 (0.0179)} & \textbf{0.5477 (0.0197)} & \textbf{0.8604 (0.0131)} & \textbf{0.5132 (0.0273)} \\
		\bottomrule[1.0pt]
	\end{tabular}}
	\label{table:pointwise_result}
\end{table*}
\subsubsection{Baselines}
We compare our \ourmethod with \revision{five} standard multi-dimensional anomaly detection methods that do not take temporal dependencies into consideration and \revision{six} recently proposed frameworks designed specifically for multivariate time-series anomaly detection. Baseline descriptions and implementation details are provided in the Appendix.

\revision{The five standard multi-dimensional anomaly detection methods are \text{Raw Signal}~\cite{garg2021evaluation}, \textbf{PCA}, \textbf{AutoEncoder}, \textbf{Kmeans} and \textbf{DAGMM}~\cite{zong2018deep}. \textbf{Raw Signal} is a simple baseline model that reconstructs any signal to zero, resulting in an error equivalent to the normalized signals themselves. Using the normalized signals, a Gaussian scoring function is employed to compute the negative log-likelihood of observing these signal values at each timestamp. This model provides insights into the nature and difficulty of the benchmark dataset.}

\revision{The six state-of-the-art frameworks for multivariate time-series anomaly detection are \textbf{LSTM-VAE}~\cite{park2018multimodal}, \textbf{OmniAnomaly}~\cite{su2019robust}, \textbf{USAD}~\cite{audibert2020usad}, \textbf{MTAD-GAT}~\cite{zhao2020multivariate}, \textbf{GDN}~\cite{deng2021graph} and \textbf{InterFusion}~\cite{li2021multivariate}.  Notably, \textbf{InterFusion}, an extension of \textbf{OmniAnomaly}, is the state-of-the-art RNN framework, while \textbf{MTAD-GAT} and \textbf{GDN} are the state-of-the-art GNN baselines for the multivariate time-series anomaly detection task.}

\subsubsection{Parameter Settings}  We train our model for 20 epochs with a batch size of 64, Adam optimizer is applied to optimize \ourmethod with learning rate of $3 \times 10^{-4}$ and $(\beta_{1},\beta_{2}) = (0.9,0.999)$. Following previous works \cite{su2019robust,li2021multivariate}, validation set ratio for SWaT, WADI and SMD are 0.1, 0.1 and 0.3 respectively. We set sliding window length, $w$, to be 5, 5, and 100 for SWaT, WADI and SMD as suggested by the original papers \cite{su2019robust,deng2021graph}. We define the hyperparameter search space as shown in Appendix, and select the hyperparameters that achieve lowest average root-mean-square error in the validation set. After hyperparameter search, the MTCL module has neighbour size, $k$, set to be 15, 30 and 10 for SWaT, WADI and SMD respectively. Across all datasets, the correlation learning module has a node dimension of 256, the retain ratio of 0.1 and saturation rate of 20. The graph convolution network and the temporal convolution network modules both have 16 output dimensions. The skip connection layers all have 32 output dimensions. We use 2 graph and temporal module layers. Lastly, for the number of principal components, we set it automatically based on the number required to achieve less than 10\% sMAPE on the validation set. Computing infrastructures and empirical computational complexity for all methods are detailed in the Appendix.

\subsection{Overall Detection Performance}\label{subsec:overall_result}
As many baseline methods do not incorporate a threshold selection mechanism \cite{zong2018deep,park2018multimodal,audibert2020usad,li2021multivariate}, we compare model performances using the Receiver Operating Characteristic (ROC) and Precision-Recall Curve (PRC) Area Under the Curve (AUC) scores by treating every timestamp as an independent observation to be classified as an anomaly or not. Under this pointwise approach, a model is required to predict the occurrence of anomaly events across the entire time-series, including when they have started and ended. The closer the ROC and PRC score is to 1, the better a model is at scoring and differentiating anomalous and non-anomalous time points. For comparison between 

\subsubsection{Baseline Comparison} The PRC and AUC results are summarized in Table \ref{table:pointwise_result} and we observe that: 
\begin{itemize} 
    \item \revision{\textbf{Proposed Framework.}} \ourmethod showed superior performances against all the other baselines with an average outperformance of 7.16 and 8.30 percentage points against the next best baseline for the ROC and PRC scores respectively. It also achieved high performance with relatively low variability and, in the case of WADI's PRC values, the performance gain is greater than 45\% when compared to the next best result. The experimental result in Table \ref{table:pointwise_result} demonstrates \ourmethod's superior performance in providing a representative anomaly indicator to inform and alert system operators. It also aids experts in deciding on an appropriate threshold for human intervention as the anomaly scores for anomalous and non-anomalous timepoints are well separated. 
    \item \revision{\textbf{Temporal Dependency.}} On average, baseline methods that consider temporal information achieve higher ROC and PRC results, validating that temporal information is paramount for detecting anomalies in multivariate time-series. The importance of effective learning of temporal cues is also evident by the performance of GDN, which did not address the temporal dependencies between time-series observations directly. Despite explicitly learning spatial correlation between multivariate variable pairs, the GDN model is less effective when adapted to the unsupervised, \textit{real-time} anomaly detection task.
    
    \item \revision{\textbf{Spatial Pairwise Correlation.}} As LSTM-VAE, OmniAnomaly and USAD do not directly capture the underlying pairwise inter-dependence among the multivariate time-series variables, they performed poorer than InterFusion and \ourmethod. Similar to our framework, InterFusion directly addresses the spatial-temporal dependencies by learning dual-view latent embeddings. Nonetheless, as InterFusion's latent embedding only encapsulates spatial correlation within a global hidden state, they do not explicitly model the relationships between variable pairs. We conjuncture that successful capturing of spatial correlation dependencies requires an \textit{explicit} graphical modeling of relationships between the multivariate variables as it evidently improves the effectiveness of a time-series anomaly detection model.
\end{itemize}
\subsubsection{Ablation Study}\label{subsec:exp_ablation_study} 
We assessed the contributions of various \ourmethod modules to anomaly detection in a targeted ablation study on SWaT and WADI. Different variants of \ourmethod were implemented with modifications across multiple modules:
\begin{itemize} 
    \item \revision{\textbf{w/o MTCL:}} \ourmethod without Multivariate time-series Correlation Learning. We replace the learned adjacency matrix, $\mathbf{A}$, with a complete digraph adjacency matrix and remove MTCL. 
    \item \revision{\textbf{w/o GCN:}} \ourmethod without the Graph Convolution Network. We remove the GCN module (including MTCL) and replace it with a linear layer.
    \item \revision{\textbf{mod. TCN:}} \ourmethod with modified Temporal Convolution Network. We modify TCN to nullify its ability in capturing multi-granular temporal clues by replacing the multi-convolution filters with a single 1x1 filter. 
    \item \revision{\textbf{w/o PCA:} \ourmethod without the PCA-based anomaly scoring module. We replace our the PCA module with standard Gaussian scoring function \cite{garg2021evaluation}. The Gaussian scoring function would correspond to the Raw Signal in Table \ref{table:pointwise_result}, but the input for this function is the forecast error from the STGNN in \ourmethod.
    \item \revision{\textbf{w/o STGNN:} \ourmethod without the STGNN. This is equivalent to the PCA model in Table \ref{table:pointwise_result}.}}
\end{itemize}
\begin{table}[!hbt]
	\small
	\caption{\centering Ablation Study - Average AUC performance\\ ($\pm$ standard deviation)}
	\resizebox{0.49\textwidth}{!}{%
	\begin{tabular}{p{55 pt}<{\centering}|p{59 pt}<{\centering}p{59 pt}<{\centering}|p{59 pt}<{\centering}p{59 pt}<{\centering}}     
	\toprule[1.0pt]

		  & \multicolumn{2}{c|}{\textbf{SWaT}} & \multicolumn{2}{c}{\textbf{WADI}}  \\ 
		  & {ROC} & {PRC} & {ROC} & {PRC} 
		    \\ 
		\cmidrule{1-5}
		\textbf{\ourmethod} & \textbf{0.8520 (0.0022)} &  \textbf{0.7628 (0.0032)} & \textbf{0.8283 (0.0179)} & \textbf{0.5477 (0.0197)} \\ \cmidrule{1-5}
		\textbf{w/o MTCL}         & 0.8457 (0.0181) & 0.7218 (0.0240) & 0.7832 (0.0046) & 0.4739 (0.0145)  \\
	    \textbf{w/o GCN}     & 0.8401 (0.0037) & 0.6800 (0.0459) & 0.7854 (0.0064) & 0.4688 (0.0103) \\
		\textbf{mod. TCN} & 0.8446 (0.0049) & 0.7324 (0.0124) & 0.7849 (0.0104) & 0.4886 (0.0148) \\
  \revision{\textbf{w/o PCA}} & \revision{0.8610 (0.0092)} & \revision{0.7509 (0.0152)} & \revision{0.7017 (0.0375)} & \revision{0.4105 (0.0625)} \\
  \revision{\textbf{w/o STGNN}} & \revision{0.8257 (0.0000)} & \revision{0.7298 (0.0000)} & \revision{0.5597 (0.0000)} & \revision{0.2731 (0.0000)} \\
		\bottomrule[1.0pt]
	\end{tabular}}
	\label{table:ablation}
\end{table}
\vspace{-0.5mm}
\revision{Focusing on MTCL module, we see a drop in performance this module is removed (\textbf{w/o MTCL}), and a a complete digraph adjacency matrix is used for modelling interactions between variables. Importantly, the degradation of performance is notably more pronounced for the WADI dataset.} We hypothesize that the noise from unimportant neighbouring nodes is more pronounced when the GCN propagate information among the variables under the WADI with 127 number of multivariate variables, as compared to SWaT that only has 51. 

\revision{Next, we scrutinize the effects of modifications to the STGNN.} We observe that the exclusion of the GCN module (\textbf{w/o GCN}) significantly degrades
the anomaly detection results. \revision{This is consistent with previous studies~\cite{li2021multivariate,deng2021graph} as modelling} of the pairwise correlations among variables can enable information flow among the interdependent univariate variable nodes, \revision{thereby improving the performance of detecting anomaly events}. Conversely, when we modify the TCN (\textbf{mod. TCN)} within \ourmethod, it also leads to decline in performance. \revision{This can be attributed to the fact that the} temporal dependency of the multivariate time-series data is less effectively captured. \revision{When the PCA-based anomaly scoring module is replaced by a Gaussian scoring function~\cite{garg2021evaluation} (\textbf{w/o PCA}), we note a reduction in performance in the WADI dataset. This performance drop can be attributed to the Gaussian function's lack of robust denoising capabilities, an area where PCA excels. Despite this, the \ourmethod still outperforms the established baselines.} 

\revision{Finally, the removal of the STGNN (\textbf{w/o STGNN}), leaving only PCA model, significantly reduces performance. This underscores the crucial role of the STGNN. While PCA can lessen inherent noise for improved representation, it is the capacity of the STGNN to recognize spatial-temporal patterns that forms a comprehensive solution for anomaly detection.}

\subsubsection{Automatic Thresholding Mechanism} \label{subsec:exp_threshold}
Our framework incorporates an automatic thresholding mechanism where the maximum anomaly score in the validation set is taken as the threshold without a need for human experts in determining the optimal threshold. Table \ref{table:auto_thres} shows the best F1 score achieved through an enumerative search of global optimal threshold against the F1 score of our automatic thresholding mechanism.

\begin{table}[!hbt]
	\centering
	\caption{F1 score of Automatic Threshold vs. Optimal Threshold.}
	\begin{tabular}{@{}c|c|c@{}}
		\toprule
		\textbf{Dataset} &  \textbf{Automatic} & \textbf{Optimal} \\ \midrule
		\textbf{SWaT}   & 0.7486 (0.0071)   &   0.7529 (0.0014)                     \\
		\textbf{WADI}    & 0.4927 (0.0487)  &  0.5711 (0.0112)   \\
		\textbf{SMD}      &   0.4065 (0.0283)    & 0.5225 (0.0011) \\ 
		\bottomrule
	\end{tabular}
	\label{table:auto_thres}
\end{table} 
\revision{The non-parametric threshold selection of \ourmethod, despite its simplicity, effectively captures the spatial-temporal dependencies of multivariate time-series during the normal period (i.e., training set). This capability allows for a notable degree of separation between anomalous and non-anomalous timepoints in the test set, as evidenced by promising F1 scores. However, it is important to note that the effectiveness of the automatic threshold is most pronounced on the SWaT dataset, and exhibits some performance drops on WADI and SMD. Moving forward, we aim to refine the thresholding process to close the gap between the automatically determined threshold and the threshold determined using best-F1 scores across a broader range of scenarios.} 

\begin{table*}[!hbt]
	\small
	\centering
	\caption{SWaT - Best F1 results at different delays. \\The best and second best performing method in each experiment is in bold and underlined respectively.}
	\scalebox{0.9}{\begin{tabular}{ p{70 pt}<{\centering}|p{40 pt}<{\centering}p{40 pt}<{\centering}p{40 pt}<{\centering}p{40 pt}<{\centering}p{40 pt}<{\centering}p{40 pt}<{\centering}p{40 pt}<{\centering}}     
		\toprule[1.0pt]
		Methods & {No delay} & {1 min} & {5 min} & {10 min}  & {20 min} & {30 min} & {60 min} \\
		\cmidrule{1-8}
          \revision{Raw Signal}            & \revision{0.2444} & \revision{0.2661} & \revision{0.2713} & \revision{0.2729} & \revision{0.2911} & \revision{0.2967} & \revision{0.7652} \\
		PCA            & 0.3049 & 0.3264 & 0.3608 & 0.4610 & 0.4684 & 0.4684 & 0.6874\\		
		AutoEncoder             & 0.3219 & 0.3264 & 0.3369 & 0.3453 & 0.3453 & 0.7473 & 0.7750 \\
		Kmeans & 0.0557 & 0.0689 & 0.0864 & 0.1062 & 0.1280 & 0.1443 & 0.4751 \\
		DAGMM     & 0.3738 & 0.3969 & 0.4120 & 0.4152 & 0.4578 & 0.6822 & 0.6822\\
		LSTM-VAE    & 0.3048 & 0.3088 & 0.3100 & 0.3100 & 0.3144 & 0.3194 & 0.5604\\
		OmniAnomaly    & 0.3200 & 0.3217 & 0.3234 & 0.3256 & 0.3295 & 0.3334 & 0.5800 \\
		USAD   & 0.3072 & 0.3169 & 0.3218 & 0.3294 & 0.3382 & 0.5208 & 0.5240 \\
   \revision{MTAD-GAT}            & \revision{0.2192} & \revision{0.2238} & \revision{0.4043} & \revision{0.4632} & \revision{0.4938} & \revision{\underline{0.7999}} & \revision{\underline{0.7999}} \\   
		GDN    & 0.3052 & 0.3125 & 0.3134 & 0.3152 & 0.3162 & 0.3182 & 0.3183 \\
		InterFusion   & \underline{0.4067} & \underline{0.4803} & \underline{0.5139} & \underline{0.5406} & \underline{0.5465} & 0.6031 & 0.6031 \\
		\cmidrule{1-8}
		\ourmethod              & \textbf{0.7576} & \textbf{0.7705} & \textbf{0.7953} & \textbf{0.7972} & \textbf{0.8093} & \textbf{0.8489} & \textbf{0.8507} \\
		\bottomrule[1.0pt]
	\end{tabular}}
	\label{table:swat_delay}
\end{table*}
\begin{table*}[!hbt]
	\small
	\centering
	\caption{WADI - Best F1 results at different delays. \\The best and second best performing method in each experiment is in bold and underlined respectively.}
	\scalebox{0.9}{\begin{tabular}{ p{70 pt}<{\centering}|p{40 pt}<{\centering}p{40 pt}<{\centering}p{40 pt}<{\centering}p{40 pt}<{\centering}p{40 pt}<{\centering}p{40 pt}<{\centering}p{40 pt}<{\centering}}     
		\toprule[1.0pt]
		Methods & {No delay} & {1 min} & {5 min} & {10 min}  & {20 min} & {30 min} & {60 min} \\
		\cmidrule{1-8}    
        \revision{Raw Signal}            & \revision{0.1948} & \revision{0.2463} & \revision{0.3009} & \revision{\underline{0.5306}} & \revision{0.5306} & \revision{0.5306} & \revision{0.5306} \\
		PCA            & 0.1172 & 0.2692 & 0.2816 & 0.3144 & 0.3144 & 0.3144 & 0.3144 \\
		AutoEncoder             & 0.2307& \underline{0.3892} & \underline{0.3892} & 0.3892 & 0.3892 & 0.3892 & 0.3892 \\
		Kmeans & 0.0258 & 0.1239 & 0.3575 & 0.4814 & 0.4814 & 0.4814 & 0.4814 \\
		DAGMM     & 0.1368 & 0.2715 & 0.2715 & 0.2715 & 0.2715 & 0.4745 & 0.4745 \\
		LSTM-VAE    & 0.1300 & 0.1806 & 0.2712 & 0.2726 & 0.3549 & 0.3549 & 0.3549\\
		OmniAnomaly    & 0.2167 & 0.2957 & 0.2957 & 0.3549 & 0.3549 & 0.3549 & 0.3549 \\
		USAD   & 0.1280 & 0.2204 & 0.2714 & 0.2728 & 0.3469 & 0.3469 & 0.3469 \\
   \revision{MTAD-GAT}            & \revision{0.1212} &     \revision{0.1286} & \revision{0.1317} & \revision{0.1691} & \revision{0.4109} & \revision{0.4109} & \revision{0.4109} \\
		GDN    & 0.1454 & 0.1539 & 0.2576 & 0.2946 & 0.2946 & 0.2946 & 0.2946 \\
		InterFusion   & \underline{0.2921} & 0.3740 & 0.3754 & 0.3754 & \underline{0.5704} & \underline{0.5774} & \underline{0.5774} \\
		\cmidrule{1-8}
		\ourmethod              & \textbf{0.4190} & \textbf{0.5796} & \textbf{0.7716} & \textbf{0.7716} & \textbf{0.7716} & \textbf{0.7716} & \textbf{0.7716} \\
		\bottomrule[1.0pt]
	\end{tabular}}
	\label{table:wadi_delay}
\end{table*}

\begin{table*}[!hbt]
	\small
	\centering
	\caption{SMD - Best F1 results at different delays. \\The best and second best performing method in each experiment is in bold and underlined respectively.}
	\scalebox{0.9}{\begin{tabular}{ p{70 pt}<{\centering}|p{40 pt}<{\centering}p{40 pt}<{\centering}p{40 pt}<{\centering}p{40 pt}<{\centering}p{40 pt}<{\centering}p{40 pt}<{\centering}p{40 pt}<{\centering}}     
		\toprule[1.0pt]
		Methods & {No delay} & {1 min} & {5 min} & {10 min}  & {20 min} & {30 min} & {60 min} \\
		\cmidrule{1-8}
          \revision{Raw Signal}            & \revision{0.4127} & \revision{0.4648} & \revision{0.5381} & \revision{0.6757} & \revision{0.7834} & \revision{0.8068} & \revision{0.8965} \\
		PCA            & 0.3102 & 0.3762 & 0.4637 & 0.5828 & 0.5956 & 0.6399 & 0.7000\\		
		AutoEncoder             & 0.4401 & \underline{0.5444} & 0.6258 & 0.7279 & 0.7670 & 0.8004 & 0.8194 \\
		Kmeans & 0.1179 & 0.1472 & 0.1923 & 0.1986 & 0.2407 & 0.2407 & 0.3527 \\
		DAGMM     & 0.3304 & 0.4117 & 0.5652 & 0.6145 & 0.6779 & 0.7060 & 0.7721\\
		LSTM-VAE    & 0.3475 & 0.3639 & 0.3812 & 0.4033 & 0.5793 & 0.5989 & 0.6658 \\
	OmniAnomaly    & 0.4649 & 0.5217 & 0.6352 & \underline{0.8215} & 0.8288 & 0.8457 & \underline{0.9093} \\
		USAD   & 0.3592 & 0.4570 & \underline{0.7036} & 0.7555 & 0.7677 & 0.0892 & 0.8484 \\
    \revision{MTAD-GAT}            & \revision{0.4688} & \revision{0.5359} & \revision{0.6544} & \revision{0.7695} & \revision{\underline{0.8355}} & \revision{\underline{0.8503}} & \revision{0.9088} \\
		GDN    & \underline{0.4734} & 0.5120 & 0.5427 & 0.7191 & 0.7269 & 0.7727 & 0.8533 \\
		InterFusion   & 0.4275 & 0.4958 & 0.6309 & 0.7392 & 0.7778 & 0.8013 & 0.8954 \\
		\cmidrule{1-8}
		\ourmethod              & \textbf{0.5070} & \textbf{0.6234} & \textbf{0.7917} & \textbf{0.8420} & \textbf{0.8548} & \textbf{0.9095} & \textbf{0.9335} \\
		\bottomrule[1.0pt]
	\end{tabular}}
	\label{table:smd_delay}
\end{table*}

\subsection{Early Detection Performance}\label{subsec:exp_edpa_result} 
As time-series anomaly events usually form contiguous anomaly segments, previous works have argued that detecting anomalies within any subset of a ground truth anomaly segment is sufficient in real-world scenarios. Based on this notion, they evaluated multivariate time-series anomaly detection models using point-adjusted (PA) approach \cite{su2019robust,audibert2020usad,li2021multivariate}. Under this approach, if any timestamp in a contiguous anomaly segment with $M_{a}$ timestamps are correctly detected as an anomaly, the PA approach considers the entire anomaly segment as correctly predicted with $M_{a}$ true positives \cite{xu2018unsupervised}. However, since any detection within a contiguous anomaly segment is treated equally, \textit{the PA approach does not reward early detections in an anomalous segment} \cite{garg2021evaluation,kim2022towards,wu2021current}. Nevertheless, early detection of anomaly events is often crucial in a wide range of practical applications and a model which can detect anomaly events early will have significant value in real-world settings \cite{ahmad2017unsupervised}. 

To evaluate early detection ability of \ourmethod and baseline methods, we adopt the metric suggested by \cite{ren2019time}, where detection of contiguous anomaly segment is only treated as true positives, if and only if an anomaly point is detected correctly and its timestamp is at most $\delta$ steps after the first anomaly of the contiguous anomaly segment. For example, $\delta = 0$ would equate to identifying an anomaly segment as early as possible without any delays and $\delta = 60$ for a time-series with second-interval would equate to detecting anomaly segment within a minute after the first anomaly timestamp. As $\delta$ becomes sufficiently large (i.e., the delay constraint is removed), the results of the early detection PA approach will be the same as the original PA approach. 

In this work, we evaluate models' early detection ability with delay 0, 1, 5, 10, 20, 30 and 60 minutes. Following previous work \cite{su2019robust,audibert2020usad,li2021multivariate} in computing model's anomaly scoring ability, we report the best F1 score for each delay. Based on Table \ref{table:swat_delay}, \ref{table:wadi_delay} and \ref{table:smd_delay}, we observe the following: 

\begin{itemize} 
	\item \revision{\textbf{Immediate Detection.}} \ourmethod showed a substantial advantage against the next baseline when $\delta = 0$ where the performance improvement is 86.27\%, 43.44\% and 2.75\% for SWaT, WADI and SMD respectively. This indicates that our model significantly outperform the simple and state-of-the-art baselines in early detection of multivariate time-series anomaly events.  
	\item \revision{\textbf{Practicality.}} On all three benchmark datasets, our proposed framework performed best across all delays, $\delta$, with the exception of 5 and 10 minutes under the SMD dataset. While model performance gaps decrease as $\delta$ is increased, our framework remains state-of-the-art even when delay is set to be 60 minutes. These results suggest that our anomaly detection model has the greatest ability at detecting not only anomalous events that require immediate attention but also anomalous events that is less hurried. \ourmethod can thus be potentially applied across a wide-range of practical applications and is dependable in a diverse range of real-world operational requirements. 
	\item \revision{\textbf{Baseline Comparison.}} Consistently, InterFusion, which learns temporal and inter-variable dependencies, ranked second in early anomaly detection. Other baselines, not addressing both dependencies, showed greater variability across delays and datasets. This validates the importance of learning both temporal and inter-dependence for model generalization across tasks and datasets.
\end{itemize}

\subsection{\revision{Root Cause Anomaly Diagnosis}}

\begin{table}[!hbt]
	\centering
	\caption{\revision{Performance on root cause diagnosis. \\Average RC-Top3 ($\pm$ standard deviation).}}
	\begin{tabular}{@{}c|c|c|c@{}}
		\toprule
		\revision{\textbf{Methods}} &  \revision{\textbf{SWaT}} & \revision{\textbf{WaDI}} & \revision{\textbf{SMD}} \\ \midrule
  \revision{\textbf{PCA}}    & \revision{0.3714 (0.0000)}  & \revision{0.3846 (0.0000)} & \revision{0.6960 (0.0000)}   \\
		\revision{\textbf{AutoEncoder}}      &   \revision{0.3543 (0.0433)}  & \revision{0.5230 (0.0644)} & \revision{0.7376 (0.0226)} \\
  	\revision{\textbf{Kmeans}}      & \revision{0.3714 (0.0000)}    &  \revision{0.4615 (0.0000)} & \revision{0.6475 (0.0000)} \\
   \revision{\textbf{DAGMM}}      &   \revision{0.1429 (0.0535)} & \revision{0.1538 (0.0000)} & \revision{0.4933 (0.0681)} \\
   \revision{\textbf{LSTM-VAE} }     &   \revision{0.3143 (0.0626)}  &  \revision{0.4615 (0.1288)} & \revision{0.5186 (0.0136)} \\
   \revision{\textbf{OmniAnomaly}}      &  \revision{ 0.3714 (0.0000)}  &  \revision{0.4768 (0.0344)} & \revision{\textbf{0.8608 (0.0044)}} \\
   \revision{\textbf{USAD}}      &  \revision{0.0400 (0.0139)}   & \revision{0.1846 (0.0377)} & \revision{0.4380 (0.0225)}\\
   \revision{\textbf{MTAD-GAT}}      & \revision{0.3829 (0.0433)} &  \revision{0.5385 (0.0543)} & \revision{0.6894 (0.0093)} \\
   \revision{\textbf{GDN}}      &  \revision{0.3047 (0.0719)} & \revision{0.4923 (0.0421)} & \revision{0.7307 (0.0444)} \\
   \midrule
   \revision{\textbf{InterFusion}}     & \revision{0.3086 (0.0313)} 
   & \revision{\underline{0.5846 (0.0422)}}  &\revision{ 0.4568 (0.0169)} \\
   \revision{\textbf{$+$ MCMC}}      &   \revision{0.2171 (0.0433)}  & \revision{0.5077 (0.0421)}  & \revision{0.7747 (0.0132)} \\
   \midrule
   \revision{\textbf{Raw Signal} }  &   \revision{0.1143 (0.0000)} & \revision{0.4615 (0.0000)} &  \revision{0.8107 (0.0000)}                    \\
  \revision{\textbf{$+$ MTCL-Graph}} & \revision{0.1200 (0.0114)} & \revision{0.5231 (0.0576)} & \revision{0.6723 (0.0460)} \\
  \midrule
  \revision{\textbf{CST-GL }}   &  \revision{\underline{0.4286 (0.0452)}}  &   \revision{0.4307 (0.0377)}    & \revision{\underline{0.8532 (0.0151)}} \\ 
  \revision{\textbf{$+$ MTCL-Graph}} & \revision{\textbf{0.4914 (0.0313)}} & \revision{\textbf{0.6154 (0.0308)}} & \revision{0.7776 (0.0430)} \\  
		\bottomrule
	\end{tabular}
	\label{table:root_cause}
\end{table}

In accordance with the approach suggested by Garg et al. \cite{garg2021evaluation}, we gauge the anomaly diagnosis performance of all models using the Root-cause top 3 metric (RC-Top3). The RC-Top3 measures instances where at least one of the genuine causes is identified among the top three causes as determined by the detection model. We provide the mean performance along with its standard deviation. Since InterFusion utilizes MCMC imputation on the original reconstruction to diagnose root causes, we present results both with and without MCMC imputation. As argued by the original authors \cite{li2021multivariate}, anomalous node variables have the potential to introduce bias into the learned embeddings within their network module, creating undesirable noise. MCMC imputation can help to dampen this noise, similar to the role of the PCA-based scorer in \ourmethod. 

For \ourmethod, we report the diagnosis performance using the ranking derived from the PCA-based method, as well as the ranking obtained after aggregating the anomaly scores from its one-hop neighbour (\ourmethod+MTCL-Graph). The latter approach leverages the learned relationships between variables from the MTCL module to diagnose root causes. This approach considers that anomalous behavior exhibited by some variables may merely be symptomatic, while the root cause could be attributed to closely related variables. To further assess the benefits of MTCL, we also present the diagnostic results of the raw signal using the MTCL-Graph (Raw Signal+MTCL-Graph). This serves to evaluate the effectiveness of MTCL in facilitating the diagnosis of anomalies, even when the raw signal alone is used.

\begin{figure*}[!hbt]
    \centering
        \includegraphics[width=0.96\textwidth]{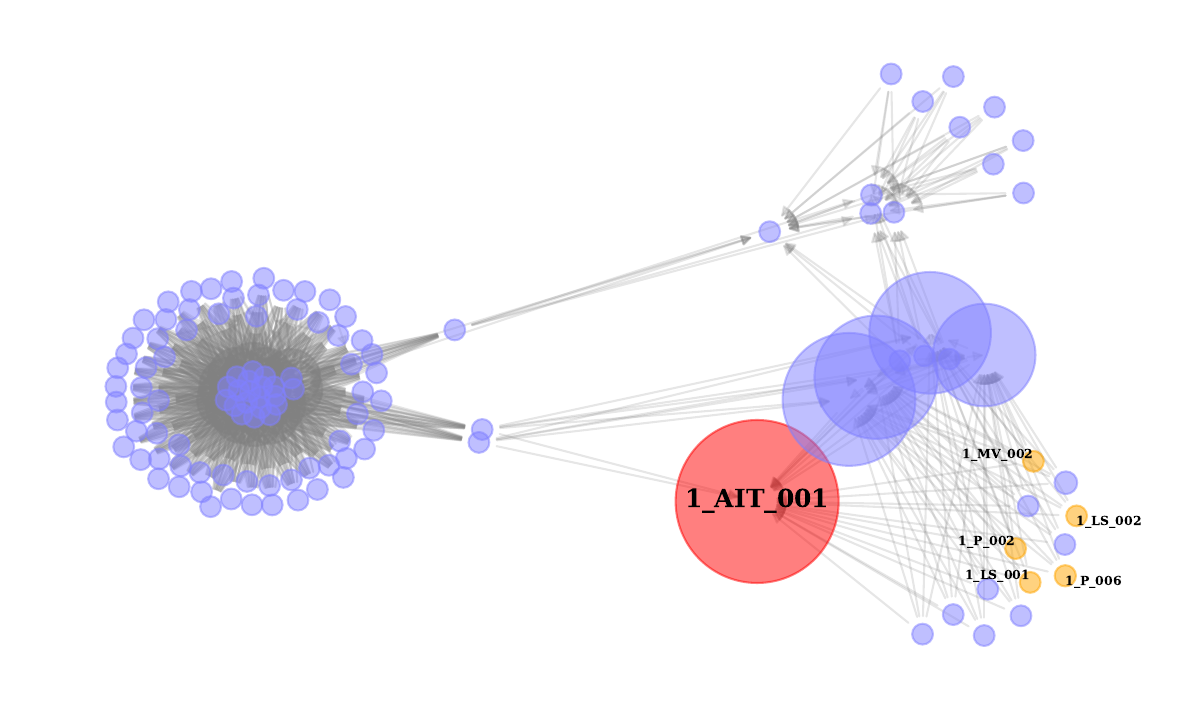}\\
       {\caption{Root cause analysis of stealth attack on Water Distribution System. Size of nodes represents the computed ranking of nodes as described in section \ref{subsubsec:rootcause_anomaly}. Red node represents the highest ranked sensor and orange nodes represent the top 5 sensors that contributed the majority of anomaly score during the first 5 minutes of attack. \revision{Apart from the highest-ranked red node, four blue nodes indicate other potential sources of the anomaly, with their sizes signifying relative importance. However, the primary node associated with the attack is the first red one, not the four blue nodes. The directed arrows represent the uni-directional relationships that \ourmethod has learned using the MTCL module and correspond to spatial dependencies between different sensors.} }\label{fig. root_cause}}
\end{figure*}
\revision{As evidenced in Table \ref{table:root_cause}, \ourmethod{} excels in identifying root causes of anomalies on SWaT and WADI, using MTCL-Graph. It comes a close second to OmniAnomaly on SMD without MTCL-Graph. As detailed in Section \ref{subsubsec:rootcause_anomaly}, the decision to diagnose root cause based directly on error contribution or one-hop distance neighbors using MTCL-Graph depends on the anomaly characteristics. SWaT and WADI often have symptomatic variables that exhibit abnormal behaviours but not causative \cite{deng2021graph,garg2021evaluation}. These variables were influenced by true root causes that exhibit normal behaviours. Thus, unearthing the true root causes requires the explicit identification of variables closely associated with symptomatic variables. This is effectively achieved with our MTCL module. }

\revision{Contrastingly, in SMD, variables that display abnormal behaviors are indeed the root causes themselves. As such, the MTCL-Graph does not provide any added benefit; instead, it is more advantageous to directly evaluate the anomaly score from the PCA-based scorer. These observations align with the diagnosis results of InterFusion. When MCMC imputation is employed, the effects of anomalous noise are mitigated, enabling InterFusion to accurately diagnose the root causes in SMD. However, MCMC imputation does not enhance results in SWaT and WADI, as it reduces the impacts of abnormal behaviors transferred to related variables. Consequently, MCMC imputation softens the anomalous effects on the actual root cause variables that do not themselves exhibit abnormal behaviors. To consolidate, we show MTCL-Graph to improve the ability in diagnosing the root cause directly from raw signals alone in SWaT and WADI, but not in SMD.} 

\revision{On the whole, \ourmethod with the aid of MTCL-Graph provides a comprehensive and actionable tool for for operators in detecting and diagnosing anomaly events.}

\subsection{Case Study in Practice}\label{subsec:exp_case_study}

To showcase \ourmethod's implementation under real-world scenarios, we conduct time-series anomaly detection case studies on WADI's Water Distribution System where the root cause of the anomaly event is known. 
\paragraph{Background} The water distribution process in WADI is segmented into three sub-processes: P1, P2 and P3. P1 involves water intake and water quality management, P2 takes in water from P1 and supplies it to the consumers and P3 returns excess water back to P1. To monitor and automate the system effectively, 127 sensors are installed. The sensors within each sub-process are intimately linked to monitor and automate the water distribution sub-process. Nonetheless, any attacks on a single sub-process will have a cascading effect on the entire water distribution system. In this experimental setting, \ourmethod is required to detect malicious attacks by ill-intentioned parties that have access to the system control from October 9, to October 11, 2017. \ourmethod is provided with 14 days of normal multi-sensors data from September 25, to  October 9, 2017, to train the model. No labels or information related to the attacks are given during training and \ourmethod is required to detect the anomalies in an unsupervised manner. 

\paragraph{Stealth Attack} At 10:55 a.m. on December 10, 2017, the attacker launch a 29-minute stealthy attack on WADI to drain an elevated reservoir by changing the reading seen by water quality sensor, 1\_AIT\_001 (i.e., the root cause of attack). Further, the attacker cleverly manipulates the root cause sensor to make the event undetectable. Consequently, determining the root cause of this attack is non-trivial given that the attacker has extensive knowledge about the WADI system and is deliberately hiding the root cause.  

\paragraph{Proposed Framework In Action} The following describes \ourmethod's  real-time anomaly detection mechanisms.
\begin{itemize} 
	\item \revision{\textbf{Automated Early Detection.}} Relying on the automated thresholding mechanism, \ourmethod alerted the human operators at 10:58 a.m. (less than 4 minutes after the attack) that a possible anomaly event has occurred and emergency intervention is required. During the attack period, the anomaly score remained high, continuously warning operators about the urgency of the attack event.
	\item \revision{\textbf{Root Cause Identification with Learned Relation.}} Looking at \ourmethod's system outputs, the human operators see 5 sensors from sub-process P1 to contribute the substantial majority of anomaly scores during this period: 1\_MV\_002, 1\_P\_002, 1\_P\_006, 1\_LS\_001 and 1\_LS\_002. After inspecting all 5 sensors, it is found that they are not the root cause but are merely the symptoms of this attack. Nevertheless, they are very likely to be related to the root cause of the stealthy attack event. Thus, the sensors most closely related to the five sensors are immediately ranked based on the \ourmethod's learned relation between variable pairs. The aggregated scores of one-hop distance neighbors in sub-process P1 ranks 1\_AIT\_001 as the sensor most associated with the 5 aforementioned sensors, as illustrated in Figure \ref{fig. root_cause}. The root cause of the attack is thus successfully identified after inspecting merely 6 out of 127 sensors. 
	\item \revision{\textbf{Informativeness with Pointwise Detection.}} The original WADI dataset assumes no human intervention and the 29-minute stealth attack ended at 11:24 a.m. Consistently, \ourmethod confirms by 11:26 a.m. that the attack has ceased, exhibiting a lag of less than two minutes. This not only allows system operator to make decisions with informed knowledge but also direct efforts on exploring the data within the most relevant time frame to thoroughly understand the anomaly events that have already occurred. 
\end{itemize}

In another WADI anomaly event, a flow sensor, 1\_FIT\_001, is attacked via false readings. To detect the root cause of this attack is again non-trivial because the false readings are within the normal range of this sensor \cite{deng2021graph}. Following the implementation above, \ourmethod is able to alert system operators that an anomalous event has occurred after just \textit{10 seconds} of the attack. Similarly, the top sensors that contributed to the anomaly score are again found not to be the the root cause. Through aggregated scores of the learned relations between sensors, \ourmethod ranked the root cause sensor, 1\_FIT\_001, as the third most possible sensor to be the root cause, correctly identifying the root cause after inspecting 8 out of 127 sensors. Lastly, \ourmethod informs that the anomaly event has ended within the precision of $\pm$1 minute.

\paragraph{Summary}  The case studies demonstrate \ourmethod's ability in (1) detecting anomalous event early, (2) significantly reducing the search range for human operators to identify the root cause by localizing the relevant variables and (3) informing operators about the duration of anomaly events with reasonable precision. Importantly, it also illustrates that joint learning of spatial-temporal and pairwise correlation relational dependencies can help a multivariate time-series anomaly detection model to detect and diagnose anomaly events, significantly reduce the destructive impact of such events on industrial systems.  

\minorrevision{Moving forward, we intend to enhance CST-GL's robustness and real-world applicability. As noted, anomalies can bias the modeling process in graph neural networks. This challenge becomes even more pronounced when anomalies introduce a long chain of anomalous effects that propagate throughout large-scale sensing networks, which are common in practice. Employing multiple GNN layers for k-hop neighbor aggregation to address such long-term dependencies might expose the system to issues like over-smoothing \cite{li2018deeper} and over-squashing \cite{topping2022understanding}. While the incorporation of the gating mechanism in Equation \ref{eq: spatial convolution} of CST-GL could alleviate the aforementioned issues, we plan to fortify our framework further by incorporating recent advances in the GNN field. Specifically, we aim to explore GNN techniques known for their stability \cite{tortorella2022leave} and consider the potentiality of including online adaptive rewiring for message passing.} 

\minorrevision{In addition, we aim to} \revision{enhance CST-GL for a broader range of applications by addressing the issues of concept drift and missing values. In terms of concept drift, we plan to implement mechanisms that can detect and quantify the magnitude of data drift, thus facilitating necessary adjustments to the model in line with evolving data distributions~\cite{webb2016characterizing,goldenberg2020pca}. For handling missing values, we intend to assess the robustness of \ourmethod by employing standard interpolation and imputation algorithms \cite{beretta2016nearest}. Furthermore, we aspire to incorporate spatial-temporal graph controlled differential equations~\cite{jin2022multivariate}, inherently suited to scenarios involving missing values.} 

\minorrevision{}

\section{Conclusion}\label{sec:conclusion}
In this work, we proposed a novel framework for \mtsad. Our model, \ourmethod, explicitly learns pairwise correlations between variables pairs of multivariate time-series data, jointly capture spatial-temporal dependencies and effectively detect anomaly events when the behaviour of time-series data deviate from the non-anomalous patterns. Experiments on three real-world datasets showed that \ourmethod outperformed \revision{eleven} baselines in general and early detection settings. \ourmethod also enables interpretation and root cause diagnosis of anomaly events in multivariate time-series data, paving the way for STGNN-based methods to be implemented in real-world applications. In the future, we will study \revision{generalizability of \ourmethod in dynamic and missing values scenarios together with} the trustworthiness of our GNN model \cite{zhang2022trustworthy} through the perspectives of robustness and explainability. We will also look into how large language models can enhance graph learning \cite{pan2023unifying,jin2023timellm} for time series data.


%

%

%
%

\ifCLASSOPTIONcaptionsoff
  \newpage
\fi



%
%
%


\bibliography{reference}
\bibliographystyle{IEEEtran}


\newpage
\appendix
Our appendix primarily provides details of the computing infrastructures and experimental settings to ensure the reproducibility of our work. \textbf{A1. Computing Infrastructures} and \textbf{A2. Implementation of Baseline} detail the computing infrastructure and the hyperparameters of the baselines we reproduced in our work, respectively. \textbf{A3. Empirical Computational Complexity} provides information about the time complexities of the baselines and our model, \ourmethod. Lastly, \textbf{A4. \ourmethod Hyperparameter Search Space} outlines the search space that we use to set the hyperparameters of \ourmethod, based on the combination of parameters that achieve the lowest average Root-Mean-Square-Error (RMSE) in the validation set.\\

\textbf{\\A1. Computing Infrastructures}
Our proposed learning framework is implemented using PyTorch 1.7.0. The computation of F1 score, ROC and PRC is acquired by Scikit-learn. All experiments are conducted on a personal computer with Ubuntu 20.04 OS, with an NVIDIA Tesla T4 GPU, a 2.20GHz Intel Xeon CPU, and 12.7 GB RAM. For model comparison with a single and five experimental runs, we use seed 0 and 0-4 respectively.

\textbf{\\A2. Implementation of Baseline}
\begin{itemize}
    \item \textbf{Raw Signal}~\cite{garg2021evaluation} is a trivial baseline model that reconstructs any signal to zero, resulting in an error that equates to the normalized signals themselves. On the normalized signals, a Gaussian scoring function is utilized to compute the negative log-likelihood of observing these signal values in each timestamp. This baseline is reproduced using the code provided in the Github repository: \url{https://github.com/astha-chem/mvts-ano-eval}. We use the dynamic gaussian scoring function (\textbf{Gauss-D}) or the 'univar\_gaussian' option in the fit\_scores\_distribution function provided in the repository.
    \item \textbf{PCA} assigns an anomaly score for each timestamp based on reconstruction error. In particular, we fit PCA on the training data, including the validation data, to obtain the mean and eigenvectors. During real-time anomaly detection testing, we project the multi-dimensional input onto a low-dimensional space, and reconstruct them back again to find the root-mean-square reconstruction error. For the number of principal components, we set it automatically based on the number required to achieve less than 10\% sMAPE.
    \item \textbf{AutoEncoder} independently assigns an anomalous score to each observation by tracking the reconstruction error using an encoder-decoder framework. The encoder is a two-layer multilayer perceptron with the dimensions being [input\_dimension, 50 and 20], and the decoder is also a a two-layer multilayer perceptron with the dimensions [20, 50 and input\_dimension]. Similar to PCA, we train the AutoEncoder on the training data, including the validation data. During real-time anomaly detection testing, we apply AutoEncoder for computing the root-mean-square reconstruction error as anomaly scores at each timestamp.
    \item \textbf{Kmeans} treats each observation as independent points, and generate multiple clusters using the training data. To determine the number of cluster, K, we use Silhouette score and we search K from 0 to 20. During real-time anomaly detection testing, we calculate the distance between multivariate observation and the centroid of its closest corresponding cluster. The computed L2 distance is used as the anomaly score for detecting anomalies.
    \item \textbf{DAGMM}~\cite{zong2018deep} joints Autoencoders and Gaussian Mixture Model to attain anomaly score using reconstruction errors generated from a low-dimensional representation. To reproduce their results on our settings, we use the Github repository:\url{https://github.com/tnakae/DAGMM}. We set the dimensions as [20, 10, 5, 1] for the compression network, and as [5, 2] for the estimation network. We set dropout ratio as 0.5. The rest of the parameters follow the default settings. Similar to PCA and AutoEncoder, we train on the training data, including the validation data. During real-time anomaly detection testing, DAGMM predicts the energy of the observation with the more energy suggesting that it more likely to be an anomaly.
    \item \textbf{LSTM-VAE}~\cite{park2018multimodal} replaces the feed-forward neural networks in the VAE with a long short-term memory (LSTM) to capture the temporal dependency of time-series data. Nevertheless, the stochasticity of variables modeled  by VAE is without temporal dependence. To reproduce the results for LSTM-VAE, we use the code from Github repository: \url{https://github.com/lin-shuyu/VAE-LSTM-for-anomaly-detection}. The hidden dimension of the network is set as 10 and number of epoch for training as 20. The window size is set as 5, 5 and 100 for SWaT, WADI and SMD, respectively. During real-time anomaly detection testing, the anomaly score is based on reconstruction errors at each timestamp.
    \item \textbf{OmniAnomaly}~\cite{su2019robust} adopts the stochastic variable connection technique, OmniAnomaly's recurrent neural network explicitly models the temporal dependencies between stochastic variables. The anomaly score is the posterior reconstruction probability of each input. Each timestamp is classified as either anomalous or non-anomalous using the Peaks-Over-Threshold method~\cite{siffer2017anomaly}. To reproduce the results from OmniAnomaly, we use the code from Github repository: \url{https://github.com/NetManAIOps/OmniAnomaly}. Following the default hyperparameters, we set the z hidden dimension as 3, RNN hidden dimension as 500, normalizing flow layers as 20, and number of epoch for training as 20. The window size is set as 5, 5 and 100 for SWaT, WADI and SMD, respectively. During real-time anomaly detection testing, the anomaly score is based on inverse of reconstruction probability at each timestamp.
    \item
    \textbf{USAD}~\cite{audibert2020usad} is an autoencoder with encoder-decoder architecture that is trained in an adversarial manner to combine the advantages of autoencoders and adversarial training. To reproduce the results from USAD, we use the code from Github repository: \url{https://github.com/manigalati/usad}. USAD utilizes one encoder network and two decoder networks. In accordance with the default setting, all networks are three-layer multilayer perceptrons, with the hidden dimension being one-half and one-quarter of the original input dimension respectively. We train USAD over 250 epochs. The window size is set as 5, 5 and 100 for SWaT, WADI and SMD, respectively. During real-time anomaly detection testing, the anomaly score is derived from the reconstruction error at each timestamp.
    \item \textbf{MTAD-GAT}~\cite{zhao2020multivariate} is an attention-based graph neural network that implicitly learns dependence relationships between the multivariate variables by assuming a complete graph between the variables. It computes both reconstruction and forecast errors to detect anomalies. To reproduce the results from MTAD-GAT, we use the code from Github repository: \url{https://github.com/ML4ITS/mtad-gat-pytorch}. The Graph Attention Networks used to model spatial and temporal cues consist of a single layer. The initial convolution layer possesses a kernel size of 7, while the number of GRU layers is also set to one, having a hidden dimension of 150. The forecast output module is designed with three hidden layers, each with hidden dimensions of 150. In contrast, the reconstruction output module contains only one hidden layer with a hidden dimension of 150. We train the MTAD-GAT over 50 epochs with a dropout rate of 0.3. The window size is set as 5, 5 and 100 for SWaT, WADI and SMD, respectively. During real-time anomaly detection testing, the anomaly score is computed based on the reconstruction and forecast error at each timestamp.
    \item \textbf{GDN}~\cite{deng2021graph} is an attention-based graph neural network that explicitly learns dependence relationships between the multivariate variables and computes forecast errors by  leveraging these relationships as anomaly scores. To reproduce the results from GDN, we use the code from Github repository: \url{https://github.com/d-ailin/GDN}. Following the default hyperparameters for WADI (SWaT), we set the embedding vector for the graph learning module to 128 (64), the number of neighbors, k, to 30 (15), and the dimension of hidden layers to 128 (64) neurons. For SMD, we set the hyperparameters to match those of SWaT. We train GDN using 50 epochs with early stopping at 10 epochs. When calculating the deviations, the original GDN model inadvertently incorporates future information into the current timestamp by normalizing errors using the full test set's median values. To rectify this, we replace this median value with the median value from the validation set. The window size is set as 5, 5 and 100 for SWaT, WADI and SMD, respectively. During real-time anomaly detection testing, the anomaly score is determined based on the normalized forecast error at each timestamp.
    \item
    \textbf{InterFusion}~\cite{li2021multivariate} explicitly learns a low-dimensional that captures inter-metric (i.e., the relationship between each univariate variable) and temporal dependency for a sequence of multivariate time-series. The anomalous score is the reconstruction probability. To reproduce the results from InterFusion, we use the code from Github repository: \url{https://github.com/zhhlee/InterFusion}. As the repository contain the parameters for each of the setting we used in this study and each setting uses a different configuration, we refer the readers to the repository for details of hyperparameters. During real-time anomaly detection testing, the anomaly score is based on inverse of reconstruction probability at each timestamp.
\end{itemize}

\textbf{\\A3. Empirical Computational Complexity\\}
The table below details the time complexities of all the models. Simple baselines, namely RawSignal, PCA and Kmeans, have negligible implementation time and are thus excluded from the table:

\begin{table}[!hbt]
	\centering
	\caption{Average training time per epoch and in total (Epoch/Total).}
	\begin{tabular}{@{}c|c|c|c@{}}
		\toprule
		\textbf{Methods} &  \textbf{SWaT} & \textbf{WaDI} & \textbf{SMD} \\ \midrule
		\textbf{AutoEncoder}      & 0.2min/0.11hr &   0.4min/0.35hr  & 0.3min/0.16hr \\
   \textbf{DAGMM}      &  0.1min/0.24hr   & 0.2min/0.79hr & 0.1min/0.05hr\\
   \textbf{LSTM-VAE}      &  0.9min/0.31hr   & 2.2min/0.72hr & 1.1min/0.53hr \\
   \textbf{OmniAnomaly}      & 1.3min/0.42hr     &  4.6min/1.53hr & 4.3min/1.43hr \\
   \textbf{USAD}      &  0.1min/0.09hr   &  0.2min/0.33hr & 0.1min/0.02hr\\
   \textbf{MTAD-GAT}      &  0.7min/0.35hr    & 4.5min/2.25hr & 0.4min/0.23hr \\
   \textbf{GDN}       &  0.4min/0.31hr & 1.2min/0.77hr &  0.3min/0.09hr \\
   \textbf{InterFusion}      &  7.8min/1.95hr   & 15.9min/3.98hr  & 3.5min/0.96hr \\
  \midrule
  \textbf{CST-GL }  &  1.3min/0.43hr  &  4.1min/1.37hr  & 0.8min/0.26hr \\
		\bottomrule
	\end{tabular}
	\label{table:time_taken}
\end{table} 

\textbf{\\A4. \ourmethod{} Hyperparameter Search Space\\}
We define the hyperparameter search space as shown in the table below, and select the hyperparameters that achieve lowest average root-mean-square error in the validation set.

\begin{table}[htb]
\centering
    \caption{The hyperparameter options we searched through \\for \ourmethod{} on validation set. }
    \begin{tabular}{l|c}
    \toprule
    \textbf{Hyperparameter} & \textbf{Search Space}\\ \toprule
    MTCL node dimension  & 64,128,256,512\\ \cline{1-2}
     MTCL retain ratio & 0.05,0.1,0.2,0.3 \\\cline{1-2}
    MTCL saturation rate, $\alpha$  & 5,10,20,30 \\ \cline{1-2}
    MTCL neighbour size, $k$    & 10, 15, 20, 30, 50  \\ \cline{1-2}
   Number of TCN Layers  & 1,2,3 \\\cline{1-2} 
   TCN output dimension   & 16,32,64  \\ \cline{1-2}
   Number of GCN Layers   & 1,2,3   \\ \cline{1-2}
    GCN output dimension  & 16,32,64  \\
    \bottomrule
    \end{tabular} \label{tab. hyperparameter}
\end{table} 

\end{document}